\documentclass[11pt,a4paper]{article}
\usepackage[hyperref]{emnlp2020}
\usepackage{times}
\usepackage{latexsym}

\usepackage{microtype}
\usepackage{booktabs}
\usepackage{amsmath}
\usepackage{amssymb}
\usepackage{xspace}
\usepackage{multirow}
\usepackage{graphicx}
\usepackage{bm}
\usepackage[ruled,vlined]{algorithm2e}
\usepackage[leftmargin=1em]{quoting}
\usepackage{titlesec}

\usepackage[T2A,T1]{fontenc}
\usepackage[utf8]{inputenc}
\usepackage[russian,english]{babel}
\usepackage{textgreek}
\usepackage{CJKutf8}

\usepackage{tikz}
\usetikzlibrary{shapes.geometric}
\usetikzlibrary{patterns}
\usepackage{pgfplots}
\usetikzlibrary{backgrounds}
\usetikzlibrary{matrix,calc}
\usepackage[dvipsnames]{xcolor}

\titlespacing*{\section}
{0pt}{1.0ex plus .2ex minus .2ex}{.8ex plus .2ex minus .2ex}
\titlespacing*{\subsection}
{0pt}{.8ex plus .0ex minus .2ex}{.6ex plus .0ex minus .2ex}
\tikzset{fit to page/.style={fit only width,fit only height},
         fit only width/.style={
                    trim left={($(current bounding box.center)-(0.5*\columnwidth,0)$)},
                    trim right={($(current bounding box.center)+(0.5*\columnwidth,0)$)},
          },
          fit only height/.style={
                     execute at end picture={%
                     \useasboundingbox let 
                           \p1=(current bounding box.north east),
                           \p2=(current bounding box.south west) in
                           \pgfextra{\pgfresetboundingbox}
                           (\x2,-0.5*\textheight) rectangle (\x1-(0,0.5*\textheight);
                    }
          }
}

\def\showcomments{1}
\if\showcomments1
\newcommand{\an}[1]{\textcolor{blue}{\small [#1 --AA]}}
\newcommand{\jun}[1]{\textcolor{purple}{\bf\small [#1 --JUN]}}
\newcommand{\hb}[1]{\textcolor{olive}{\bf\small [#1 --Haibo]}}
\newcommand{\zj}[1]{\textcolor{brown}{\bf\small [#1 --ZJ]}}
\newcommand{\gn}[1]{\textcolor{orange}{\bf\small [#1 --GN]}}
\else 
\newcommand{\an}[1]{}
\newcommand{\jun}[1]{}
\newcommand{\hb}[1]{}
\newcommand{\zj}[1]{}
\newcommand{\gn}[1]{}
\fi

\DeclareMathOperator*{\argmax}{argmax}
\DeclareMathOperator*{\argmin}{argmin}
\def\smallcol{\hskip 6pt}
\def\tinycol{\hskip 4pt}
\def\mask{\text{$\langle$mask$\rangle$}\xspace}
\def\xph{\text{[X]}\xspace}
\def\yph{\text{[Y]}\xspace}
\def\blank{\_\xspace}
\def\langnum{23\xspace}
\def\methodabbr{X-FACTR\xspace}

\aclfinalcopy 
\def\aclpaperid{2630} 

\newcommand\blfootnote[1]{%
  \begingroup
  \renewcommand\thefootnote{}\footnote{\noindent #1}%
  \addtocounter{footnote}{-1}%
  \endgroup
}

\title{\methodabbr: Multilingual Factual Knowledge Retrieval\\ from Pretrained Language Models}

\author{
Zhengbao Jiang$^\dag$, Antonios Anastasopoulos$^{\clubsuit,*}$, Jun Araki$^\ddag$, Haibo Ding$^\ddag$, Graham Neubig$^\dag$  \\
  $^\dag$Languages Technologies Institute, Carnegie Mellon University\\
  $^\clubsuit$Department of Computer Science, George Mason University\\
  $^\ddag$Bosch Research}

\date{}

\begin{document}
\abovedisplayskip=3pt
\abovedisplayshortskip=3pt
\belowdisplayskip=3pt
\belowdisplayshortskip=3pt
\maketitle
\begin{abstract}
Language models (LMs) have proven surprisingly successful at capturing factual knowledge by completing cloze-style fill-in-the-blank questions such as ``Punta Cana is located in \_.''
However, while knowledge is both written and queried in many languages, studies on LMs' factual representation ability have almost invariably been performed on English.
To assess factual knowledge retrieval in LMs in different languages, we create a multilingual benchmark of cloze-style probes for \langnum typologically diverse languages.
To properly handle language variations,
we expand probing methods from single- to multi-word entities, and develop several decoding algorithms to generate multi-token predictions.
Extensive experimental results provide insights about how well (or poorly) current state-of-the-art LMs perform at this task in languages with more or fewer available resources.
We further propose a code-switching-based method to improve the ability of multilingual LMs to access knowledge, and verify its effectiveness on several benchmark languages.
Benchmark data and code have been released at \url{https://x-factr.github.io}.\blfootnote{$*$: Work done at Carnegie Mellon University. The first two authors contributed equally.}
\end{abstract}

\section{Introduction}

Language models (LMs; \cite{church-1988-stochastic,kneser1995improved,bengio2003neural}) learn to model the probability distribution of text, and in doing so capture information about various aspects of the syntax or semantics of the language at hand.
Recent works have presented intriguing results demonstrating that modern large-scale LMs also capture a significant amount of \emph{factual knowledge} \cite{petroni-etal-2019-language,jiang-2019-lpaqa,poerner-2019-ebert}.
This knowledge is generally probed by having the LM fill in the blanks of cloze-style prompts such as ``Obama is a \blank by profession.'', where these prompts are invariably written in English.
However, it goes without saying that there are many languages of the world other than English, and it is quite conceivable that (1) users may want to query this factual knowledge in other languages, and (2) some facts will be written in non-English languages and thus multilingually trained LMs (hereinafter, M-LMs) may be more equipped to recall these facts in the languages of the original data.
In this paper, we study the intersection of \emph{multilinguality and the factual knowledge included in LMs}.

\begin{figure}[t]
\centering
\small
\begin{tabular}{@{\tinycol}l@{\tinycol}l@{\tinycol}}
\multicolumn{2}{l}{
\pgfplotstableread[row sep=\\,col sep=&]{
lang & size & siz \\
en& 6.1  & 6.07 \\
fr& 2.2  &	2.16 \\
nl& 2.0  &	2.0 \\
ru& 1.6  &	1.56 \\
es& 1.5 &	1.53 \\
jp& 1.2 &	1.22 \\
vi& 1.2 &	1.22 \\
zh& 1.1 &	1.12 \\
hu& 0.5 &	0.48 \\
ko& 0.5 &	0.47 \\
tr& 0.4  &	0.36 \\
he& 0.3 &	0.25 \\
}\wikidata
\def\mystrut{\vphantom{h}}

\begin{tikzpicture}[trim left=0cm,trim right=0cm]
    \begin{axis}[
            ybar,
            every axis plot post/.style={/pgf/number format/fixed},
            bar width=.23cm,
            width=8.5cm,
            height=3cm,
            ymajorgrids=true,
            yminorgrids=true,
            legend style={draw=none,at={(0.2,0.9)},anchor=west},
            xtick={en,fr,nl,ru,es,jp,vi,zh,hu,ko,tr,he},
            symbolic x coords={en,fr,nl,ru,es,jp,vi,zh,hu,ko,tr,he},
            every x tick label/.append style={font=\mystrut},
            tick pos=left,
            hide y axis,
            axis x line*=bottom,
            nodes near coords,
            nodes near coords align={vertical},
            every node near coord/.append style={font=\tiny,color=black},
            title={Wikipedia Size (in million articles)},
            title style={yshift=-.5cm},
            ymin=0,ymax=6.1,
            ylabel shift={-1cm},
            enlarge x limits=0.025,
        ]
        \addplot [style={blue,fill=blue,mark=none}] table[x=lang,y=size]{\wikidata};
    \end{axis}
\end{tikzpicture}} \\
\multicolumn{2}{l}{
\pgfplotstableread[row sep=\\,col sep=&]{
lang & size & siz \\
el& 0.2 &	0.17 \\
war& 0.2 &	0.16 \\
mr& 0.1 &	0.05 \\
mg& 0.09 &	0.09  \\
bn& 0.09 &	0.09  \\
tl& 0.07 &	0.07  \\
sw& 0.06 &	0.06  \\
pa& 0.04 &	0.04  \\
ceb& 0.03 & 0.03 \\
yo& 0.03 &	0.03  \\
ilo& 0.02 &	0.02  \\
}\wikidatatwo
\def\mystrut{\vphantom{h}}

\begin{tikzpicture}[trim left=0cm,trim right=0cm]
    \begin{axis}[
            ybar,
            every axis plot post/.style={/pgf/number format/fixed},
            bar width=.23cm,
            width=8.5cm,
            height=2cm,
            ymajorgrids=true,
            yminorgrids=true,
            legend style={draw=none,at={(0.2,0.9)},anchor=west},
            xtick={el,war,mr,mg,bn,tl,sw,pa,ceb,yo,ilo},
            symbolic x coords={el,war,mr,mg,bn,tl,sw,pa,ceb,yo,ilo},
            every x tick label/.append style={font=\mystrut},
            tick pos=left,
            hide y axis,
            axis x line*=bottom,
            nodes near coords,
            nodes near coords align={vertical},
            every node near coord/.append style={font=\tiny,color=black},
            ymin=0,ymax=6.1,
            ylabel shift={-1cm},
            enlarge x limits=0.025,
        ]
        \addplot [style={blue,fill=blue,mark=none}] table[x=lang,y=size]{\wikidatatwo};
    \end{axis}
\end{tikzpicture}} \\
\midrule
fact & $\langle$Bloomberg L.P., founded\_in, New York$\rangle$ \\
en prompt \ & \xph was founded in \yph. \\
\midrule
es prompt \ & \xph fue [fundar.Gerund;X] en \yph. \\
\hspace{0.4cm} & \hspace{0.1cm} $\downarrow$ \hspace{2.2cm} $\downarrow$ \hspace{1.0cm} $\downarrow$ \\
es sentence & Bloomberg L.P. fue fundada en \underline{\mask$\times 1\sim 5$}. \\ 
es outputs & \begin{tabular}{@{}l@{\smallcol}c@{\smallcol}r@{}} prediction & \#tokens & confidence \\\hline 2012 & 1 & -1.90 \\ \textbf{Nueva York} & 2 & -0.61 \\ EE. UU & 3 & -1.82 \\ Chicago, Estados Unidos & 4 & -3.58 \\ 2012 Bloomberg L.P & 5 & -3.06 \end{tabular} \\
\bottomrule
\end{tabular}
\caption{\methodabbr contains \langnum languages, for which the data availability varies dramatically. Prompts get instantiated to produce grammatical sentences with different numbers of mask tokens and are used to obtain predictions for \yph. In this Spanish example, the verb gerund ``fundar'' \textit{to found} is rendered as ``fundada'' to agree in gender and number with the subject ``Bloomberg L.P.''. The final prediction is in bold.}
\label{fig:example}
\end{figure}

We create a new multilingual benchmark for probing factual knowledge in LMs -- the Cross-lingual FACTual Retrieval benchmark (\methodabbr).
\methodabbr~shares a similar formulation as the LAMA benchmark of \citet{petroni-etal-2019-language}, which assesses whether LMs have memorized a fact (i.e., a subject-relation-object triple) by having LMs predict the blank (i.e. object) in a cloze-style prompt for each relation after filling in the subject.
We manually create such prompts for \langnum languages spanning different language families and different levels of data availability (\autoref{sec:languages}).
Because many languages that we handle are morphologically rich, we design a morphology-sensitive annotation schema (see example in \autoref{fig:example}) that can properly instantiate prompts using entity metadata (e.g. gender) and a morphological inflection model (\autoref{sec:prompts}).

In addition, while previous works \cite{petroni-etal-2019-language,jiang-2019-lpaqa,poerner-2019-ebert} have limited examination to single-token entities (e.g.~``France''), we expand our setting to include multi-token entities (e.g.~``United States''), which comprise more than 75\% of facts included in our underlying database (Wikidata; \autoref{sec:facts}).
We propose several decoding algorithms for prediction of these multi-token entities using masked LMs (\autoref{sec:multi_token}).
We discuss the related work in depth in \autoref{sec:related}.

We perform experiments on \methodabbr (\autoref{sec:exp}), comparing and contrasting across languages and LMs to answer the following research questions: (1) How and why does performance vary across different languages and models? (2) Can multilingual pre-training increase the amount of factual knowledge in LMs over monolingual pre-training? (3) How much does knowledge captured in different languages overlap?
We find that the factual knowledge retrieval of M-LMs in high-resource languages is easier than in low-resource languages, but the overall performance is relatively low, indicating that this is a challenging task.
We analyze the types of failure cases, shedding light on future directions to improve factual knowledge in M-LMs.
In addition, multilingual pre-training does not necessarily lead to a higher recall of facts compared to language-specific monolingual pre-training. The knowledge memorized by M-LMs in fact is largely distinct across languages, with almost 50\% of facts being recalled in only one language.

Inspired by the above observations, we propose a code-switching-based objective function to improve the ability of M-LMs to access knowledge using queries from a variety of languages.
We replace entities in a sentence from the original language with counterparts in another language, and further fine-tune the LM on these code-switched data (\autoref{sec:code_switch}).
We perform experiments on three languages (French, Russian, and Greek, code-switched with English).
Results demonstrate that this code-switching-based learning can successfully improve the knowledge retrieval ability with low-resource language prompts.

\section{Retrieving Facts from LMs}
\label{sec:lama}

In this paper we follow the protocol of \citet{petroni-etal-2019-language}'s English-language LAMA benchmark, which targets factual knowledge expressed in the form of subject-relation-object triples from Wikidata\footnote{\url{https://www.wikidata.org/}} curated in the T-REx dataset \cite{elsahar-2018-trex}.
The cloze-style prompts used therein are manually created and consist of a sequence of tokens, where \xph and \yph are placeholders for subjects and objects (e.g. ``\xph is a \yph by profession.'').
To assess the existence of a certain fact, \xph is replaced with the actual subject (e.g. ``Obama is a \mask by profession.'') and the model predicts the object in the blank
$\hat{y}_i = \argmax_{y_i} p(y_i|\bm{s}_{i:i}),$
where $\bm{s}_{i:i}$ is the sentence with the $i$-th token masked out.
Finally, the predicted fact is compared to the ground truth.
In the next section, we extend this setting to more languages and predict multiple tokens instead of a single one.

\section{Multilingual Multi-token Factual Retrieval Benchmark}
\subsection{Languages}
\label{sec:languages}
In sampling the languages to create our multilingual benchmark, we attempted to create a subset as diverse as possible with regards to data availability, typology, and script -- within the constraints of requiring inclusion in Wikidata and standard pre-trained M-LMs.
To this end, we created prompts in \langnum languages: English, French, Dutch, Spanish, Russian, Japanese, Chinese, Hungarian, Hebrew, Turkish, Korean, Vietnamese, Greek, Cebuano, Marathi, Bengali, Waray, Tagalog, Swahili, Punjabi, Malagasy, Yoruba, and Ilokano.

Our subset includes languages from~11 families (the Indo-European ones include members of the Germanic, Romance, Greek, Slavic, and Indic genera), using~10 different scripts.
Our languages display high variance with respect to Wikipedia presence, a proxy for overall data availability, ranging from very large to very small (see~\autoref{fig:example}).\footnote{We excluded bot-made pages for Cebuano and Waray.}

\begin{table*}[t]
\scriptsize
\centering
\begin{tabular}{@{}l@{\smallcol}r@{\smallcol}r@{\smallcol}r@{\smallcol}r@{\smallcol}r@{\smallcol}r@{\smallcol}r@{\smallcol}r@{\smallcol}r@{\smallcol}r@{\smallcol}r@{\smallcol}r@{\smallcol}r@{\smallcol}r@{\smallcol}r@{\smallcol}r@{\smallcol}r@{\smallcol}r@{\smallcol}r@{\smallcol}r@{\smallcol}r@{\smallcol}r@{\smallcol}r@{\smallcol}r@{\smallcol}r@{}}
\toprule
 & en & fr & nl & es & ru & ja & zh & hu & he & tr & ko & vi & el & bn & ceb & mr & war & tl & sw & pa & mg & yo & ilo \\
\midrule
\#all & 45.7 & 40.2 & 38.3 & 37.1 & 26.3 & 25.1 & 23.1 & 20.4 & 17.1 & 16.1 & 16.1 & 13.6 & 13.0 & 9.4 & 8.2 & 7.9 & 7.3 & 7.1 & 6.8 & 5.5 & 4.9 & 4.6 & 4.1 \\
\#single-token & 18.9 & 13.9 & 12.8 & 13.5 & 3.4 & 1.3 & 0.2 & 6.2 & 1.1 & 2.5 & 2.0 & 3.9 & 0.7 & 0.1 & 3.3 & 0.2 & 3.0 & 3.2 & 2.8 & 0.1 & 1.7 & 0.9 & 2.1 \\
\#multi-token & 26.8 & 26.4 & 25.5 & 23.6 & 22.9 & 23.8 & 22.9 & 14.2 & 16.0 & 13.6 & 14.1 & 9.7 & 12.3 & 9.3 & 4.9 & 7.7 & 4.4 & 3.9 & 4.0 & 5.4 & 3.2 & 3.7 & 2.0 \\
\bottomrule
\end{tabular}
\caption{\methodabbr benchmark statistics (in thousands). More details in the Appendix (\autoref{tab:detail_data_stat} and \autoref{fig:token}).
}
\label{tab:data_stat}
\end{table*}

\subsection{Facts}
\label{sec:facts}
While \citet{petroni-etal-2019-language} and follow-up works focus on entities that can be represented by a single token, since many popular entities consist of multiple tokens (e.g. ``United States''), we argue that it is crucial to include multi-token entities in the benchmark to make the evaluation unbiased.
Similar to \citet{petroni-etal-2019-language}, we use the T-REx dataset to collect facts for our benchmark.
Since T-REx aligns facts from Wikidata with sentences in abstract sections from DBpedia, we can estimate the commonality of each fact based on its frequency of being grounded to a sentence in these abstracts.

For each of the 46 relations in T-REx, we sample 1000 subject-object pairs with probability proportional to their frequency.
Frequency-proportional sampling makes the distribution of the facts in our benchmark close to real usage and covers facts of different popularity.
To keep the benchmark unbiased, we did not constrain the facts with any language-related criteria (e.g., require the entities to have translations in all languages we considered). As a result, some entities (either subjects or objects) might not have translations in all languages.
The number of facts in different languages in our multilingual multi-token \methodabbr benchmark is shown in \autoref{tab:data_stat}.
Because many modern pre-trained M-LMs almost invariably use some variety of sub-word tokenization, the number of tokens an entity contains will depend on the tokenization method used in the LM.
We report the statistics based on the WordPiece tokenization used in multilingual BERT \cite{devlin-etal-2019-bert}. The tokenization scheme statistics for the other M-LMs are similar.

\subsection{Prompts}
\label{sec:prompts}
Some languages we include in the benchmark require additional handling of the prompts to account for their grammar or morphology. For example, (some) named entities inflect for case in languages like Greek, Russian, Hebrew, or Marathi. In some languages syntactic subjects and objects need to be in particular cases. Similarly, languages often require that the verb or other parts of the sentence agree with the subject or the object on some morphological features like person, gender, or number. 

Our prompts provide the necessary information in order to generate grammatical sentences, given the gender and number of the entities. For example, the Russian prompt for ``\xph was born in \yph'' is:
\begin{quoting}
\small
\foreignlanguage{russian}{$\Big[$X.Nom$\Big]$ $\Big[$родился;X=MASC | родилась;X=FEM | родилось;X=NEUT$\Big]$ в $\Big[$Y.Ess$\Big]$}.
\end{quoting}
The prompt denotes that the subject (\xph) needs to be in the nominative (\texttt{Nom}) case and the object (\yph) needs to be inflected in the essive case (\texttt{Ess}).
The prompt also accounts for the variation of the gender of \xph providing options (separated by \texttt{|}) for the subject being masculine, feminine, or neuter (\texttt{MASC}, \texttt{FEM}, \texttt{NEUT} respectively). 

Everything within square brackets gets concretely instantiated given the subject and object.
Grammatical gender is assigned through a combination of Wikidata information and language-specific heuristics, constructed based on feedback from native speakers of each language. When the entity corresponds to a person, we retrieve their ``sex\_or\_gender'' properties from Wikidata. In addition, for languages like Greek or French, the gender of an entity can be inferred with fairly high certainty given the form of the word (e.g. looking at the ending). Last, some categories of entities (such as cities, countries, organizations, etc, which can be obtained using the ``instance\_of'' Wikidata property) often get assigned a general grammatical case based on the category.

Once all the morphological features have been specified as detailed above, we use the \texttt{unimorph\_inflect} package~\cite{anastasopoulos-neubig-2019-pushing} to generate the appropriately inflected surface form of the bracketed words.\footnote{\url{https://github.com/antonisa/unimorph_inflect}} We note that the target entity (\yph) might also need to be inflected, as in the above Russian example, in which case we require the model's predictions to match the inflected target forms.

To verify the quality of the prompts we performed user studies with native speakers, finding that 88\% on average were judged as natural and grammatically correct. Details are shown in~\autoref{app:prompt}, but it is worth noting that the majority of errors are due to prompts being awkward or incorrect for some senses captured by the relation, and not due to our gender heuristics or automatic inflection. This issue is also present in the LAMA English prompts~\cite{jiang-2019-lpaqa}.

\subsection{Evaluation}
As noted in \citet{petroni-etal-2019-language}, because some subject-relation pairs might have multiple correct objects (e.g., America maintains diplomatic relations with multiple countries), we collect all valid objects and judge a prediction as correct if it can match any object (e.g., both France and Canada are correct).
Since an entity might have multiple aliases (e.g., ``America'' and ``the US''), we collect all aliases for each entity from Wikidata, and the prediction is marked as correct if it can match any one of them after lowercasing.

\section{Multi-token Decoding}
\label{sec:multi_token}

As \autoref{tab:data_stat} shows, many facts involve multi-token entities and thus a LM would need to predict these entities in multiple steps.
Generating multiple predictions is straightforward for traditional left-to-right LMs \cite{sundermeyer-2015-lstmlm,radford-2019-gpt2}, where we can autoregressively decode the next token conditioned on previous tokens.
However, many pre-trained LMs such as BERT \citep{devlin-etal-2019-bert} are \emph{masked} LMs that predict individual words given left and right contexts, and decoding from such masked LMs remains an open problem \cite{lawrence-etal-2019-attending,salazar-2019-pll,ghazvininejad-etal-2019-mask,wang-2019-bertmouth,cho2020BERT}.
We systematically examined different multi-token decoding algorithms from three orthogonal perspectives: (1) how the initial predictions are produced, (2) how to refine the predictions, and (3) other commonly used components in neural text generation systems.
We assume that the following conditional probability distribution is defined by the masked LM for a sentence with $n$ tokens:
\begin{equation}
p(x_k | x_1^\prime, ..., x_{k-1}^\prime, \mask_k, x_{k+1}^\prime, ..., x_n^\prime),
\label{eq:cond}
\end{equation}
where the subscript of \mask indicates its position, and the surrounding token $x_{\cdot}^\prime$ can either be an actual word $x_{\cdot}$ or \mask.
We aim to handle sentences containing multiple mask tokens conditioning on the surrounding actual words:
\begin{equation}
\resizebox{0.85\hsize}{!}{
$\bm{s}_{i:j} = x_1, ..., x_{i-1}, \mask_i, ..., \mask_j, x_{j+1}, ..., x_n,$}
\label{eq:prompt}
\end{equation}
where $\bm{s}_{i:j}$ indicates a sentence with the $i$-th to $j$-th tokens masked out.\footnote{We assume that the mask tokens are consecutive for notation simplicity, although all following methods/equations can be easily adapted to non-consecutive cases.}

\subsection{Initial Prediction and Refinement}\label{sec:dec}

Given a sentence with multiple mask tokens, e.g., \autoref{eq:prompt}, we can either generate outputs in parallel independently or one at a time conditioned on the previously generated tokens.
These methods are similar to the prediction problems that BERT \cite{devlin-etal-2019-bert} and XLNet \cite{yang-2019-xlnet} perform in their pre-training stages respectively.
We define $\bm{c} \in \mathbb{R}^n$ as the probability of each prediction, with details varying by prediction methods.

After all mask tokens are replaced with the initial predictions, i.e., $\hat{\bm{s}}_{i:j} = x_1, ..., \hat{y}_i, ..., \hat{y}_j, ..., x_n$, we can further refine the predictions by iteratively modifying one token at a time until convergence or until the maximum number of iterations is reached. Here we outline the algorithms with high-level descriptions, and provide concrete details in~\autoref{ap:decoding}.

\begin{figure}
\centering
\includegraphics[width=1.0\columnwidth, clip, keepaspectratio]{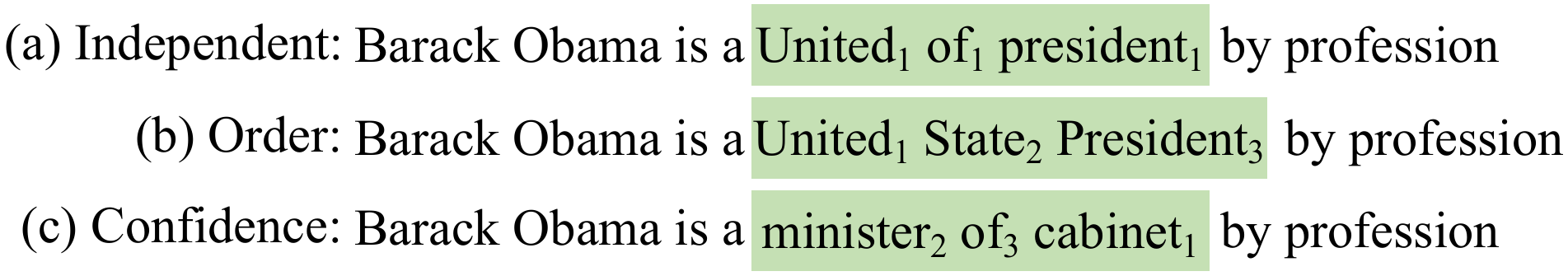}
\caption{Illustration of three initial prediction and refinement methods. Green boxes are mask tokens to be filled, and subscripts indicate the prediction order.}
\label{fig:decode}
\end{figure}

\noindent
\textbf{Independent.}
For independent initial prediction (\autoref{fig:decode}a), the mask tokens are all predicted in parallel (at once).
We also consider two autoregressive methods for initial prediction or refinement.

\noindent
\textbf{Order-based.} Mask tokens are predicted from left to right, in each step conditioning also on the previously generated tokens (\autoref{fig:decode}b). In the refinement stage, we modify predictions also from left to right, and convergence is reached when there are no changes in a left-to-right scan.

\noindent
\textbf{Confidence-based.} In each step, we choose the prediction with the highest probability, so the order of predictions can be arbitrary (\autoref{fig:decode}c).
In the refinement stage, we choose from all predicted tokens the one with the lowest confidence (i.e., the lowest probability) and re-predict it similarly to~\citet{ghazvininejad-etal-2019-mask}.
Convergence is reached when the re-predicted token is the same as the original token.

\subsection{Final Prediction}
Because we do not know the number of tokens of the ground truth in advance, we enumerate from 1 to $M$ mask tokens and choose the final prediction based on the confidence.
Given the prompt in \autoref{eq:prompt}, the simplest way to compute the confidence is pseudo log likelihood, which is the sum of log probabilities of each predicted token conditioned on the other tokens \cite{salazar-2019-pll}:
$v(j-i+1) = \sum_{k=i}^{j}{\log c_k}$, \label{eq:conf}
where $c_k$ is the confidence (probability) of the $k$-th predicted token, and $v(m)$ is the overall prediction confidence with $m$ initial mask tokens.
Among $M$ predictions, we choose the one with the highest confidence.

\subsection{Additional Components}\label{sec:dec_add}
We also investigate additional components commonly used in neural generation systems.
Specifically, we consider \textbf{length normalization} in computing the final confidence (i.e., divide $v(m)$ by the number of mask tokens $m$) because a simple sum might favor short predictions.
In addition, the confidence value $\bm{c}$ in previous methods contains probabilities when the predictions are first generated, which will become stale once the surrounding tokens change \cite{ghazvininejad-etal-2019-mask}.
We consider \textbf{re-computing confidence} $\bm{c}$ whenever a change happens.
Last, we attempted \textbf{beam search} to keep track of the most plausible $B$ predictions at each step.
Details of these components can be found in \autoref{ap:decoding}, along with a general schema of the overall decoding algorithm in \autoref{alg:overall}.

\section{\methodabbr Benchmark Performance}
\label{sec:exp}
\paragraph{Implementation Details.} We use the implementations of different multilingual/monolingual pre-trained LMs in the Transformers library \cite{wolf-2019-huggingface}.
We examine 3 multilingual pre-trained LMs, M-BERT, XLM, XLM-R \cite{devlin-etal-2019-bert,conneau-2019-xlm,conneau-2020-xlmr},\footnote{Yoruba is not in the training data of XLM and XLM-R.} and 8 monolingual pre-trained LMs, BERT (en), CamemBERT (fr), BERTje (nl), BETO (es), RuBERT (ru), Chinese BERT (zh), BERTurk (tr), and GreekBERT (el) \cite{martin-2019-camembert,vries-2019-bertje,canete-2020-beto,kuratov-2019-rubert,schweter-2020-berturk}.
Details of these models can be found in \autoref{ap:berts}.

We set the maximal number of mask tokens to $M\!=\!5$ for English, French, Dutch, and Spanish. In these languages more than~90\% of the entities are split into $\leq$5 tokens. For all other languages we use $M\!=\!10$.
This is expected because the vocabulary of M-LMs based on WordPiece tokenization is dominated by frequent words and low-resource-language words tend to split into more pieces~\cite{acs2019exploring}.
We set the maximal number of iterations to $T\!=\!2M$, so that we can approximately refine all the predicted tokens once for a sentence with $M$ mask tokens (the initial prediction takes exactly $M$ iterations).
In our main results, we report results with two decoding algorithms: the simplest independent generation method and the confidence-based method for both initial and refinement predictions.
The latter performs better than order-based methods, as we will show in~\autoref{tab:decode}.
To save computation time, we only use confidence re-computation for $M=5$.
We discuss computation complexity in \autoref{ap:decoding}.

\paragraph{Evaluation Metrics.}
We follow \citet{petroni-etal-2019-language}, computing the accuracy of predicted objects for each relation and macro-average them as final scores.
For fine-grained analysis of different decoding methods, pre-trained LMs, and languages, we report results on \textbf{all} facts as well as on subsets consisting only of single-token objects (\textbf{single}) and multi-token objects (denoted as \textbf{multi}).

\subsection{Experimental Results}
\begin{figure*}
\pgfplotstableread[row sep=\\,col sep=&]{
language & mbertdefall & mbertdefsingle & mbertdefmulti & mbertbestall & mbertbestsingle & mbertbestmulti & xlmdefall & xlmdefsingle & xlmdefmulti & xlmbestall & xlmbestsingle & xlmbestmulti & xlmrdefall & xlmrdefsingle & xlmrdefmulti & xlmrbestall & xlmrbestsingle & xlmrbestmulti & specdefall & specdefsingle & specdefmulti & specbestall & specbestsingle & specbestmulti \\
en & 13.57 & 22.40 & 5.57 & 12.00 & 12.91 & 10.08 & 9.03 & 20.74 & 4.75 & 5.30 & 8.79 & 5.63 & 8.19 & 15.21 & 3.32 & 4.43 & 5.19 & 3.86 & 17.92 & 31.21 & 5.88 & 10.53 & 19.01 & 3.44 \\
fr & 10.21 & 19.07 & 3.92 & 6.30 & 7.77 & 4.78 & 7.44 & 16.58 & 4.03 & 4.13 & 6.14 & 3.56 & 4.70 & 11.29 & 2.34 & 2.90 & 4.38 & 2.33 & 10.36 & 20.30 & 4.88 & 6.20 & 15.50 & 3.09 \\
nl & 12.42 & 25.21 & 4.42 & 8.55 & 12.20 & 5.22 & 7.53 & 18.38 & 3.00 & 4.46 & 6.48 & 4.06 & 4.42 & 10.95 & 2.58 & 2.67 & 3.57 & 2.70 & 9.84 & 19.22 & 3.40 & 5.18 & 8.21 & 3.06 \\
es & 14.30 & 24.25 & 4.90 & 7.47 & 9.13 & 5.11 & 7.40 & 16.44 & 3.40 & 3.18 & 4.18 & 3.09 & 6.50 & 13.37 & 3.29 & 4.33 & 4.93 & 4.17 & 10.94 & 19.07 & 6.10 & 6.07 & 5.22 & 6.40 \\
ru & 1.87 & 4.58 & 0.96 & 2.54 & 3.65 & 1.86 & 2.29 & 7.62 & 1.40 & 2.14 & 3.61 & 2.01 & 5.26 & 14.41 & 3.77 & 5.53 & 14.15 & 4.12 & 6.77 & 9.64 & 5.50 & 6.80 & 9.22 & 5.59 \\
ja & 0.85 & 7.13 & 0.48 & 1.51 & 6.50 & 1.21 & 5.77 & 24.95 & 3.04 & 5.95 & 18.60 & 4.24 & 2.30 & 9.23 & 2.07 & 4.41 & 8.82 & 4.21 & 0.00 & 0.00 & 0.00 & 0.00 & 0.00 & 0.00 \\
zh & 2.50 & 9.61 & 2.22 & 6.62 & 5.21 & 6.49 & 5.83 & 17.12 & 2.57 & 3.40 & 10.44 & 1.38 & 4.63 & 11.85 & 4.49 & 5.30 & 11.79 & 5.17 & 5.47 & 3.55 & 5.18 & 10.07 & 3.04 & 9.80 \\
ko & 4.08 & 21.14 & 1.61 & 4.70 & 16.15 & 2.88 & 5.33 & 13.28 & 3.24 & 5.23 & 11.15 & 3.82 & 5.11 & 16.71 & 2.61 & 5.64 & 15.88 & 3.44 & 0.00 & 0.00 & 0.00 & 0.00 & 0.00 & 0.00 \\
vi & 8.34 & 17.69 & 2.91 & 9.20 & 14.60 & 5.22 & 6.86 & 12.12 & 3.93 & 6.26 & 8.98 & 4.38 & 8.52 & 14.22 & 5.12 & 9.51 & 12.56 & 6.97 & 0.00 & 0.00 & 0.00 & 0.00 & 0.00 & 0.00 \\
el & 4.46 & 21.11 & 2.16 & 6.77 & 13.69 & 5.72 & 7.10 & 18.03 & 5.16 & 7.56 & 13.86 & 6.50 & 6.28 & 27.33 & 2.94 & 7.25 & 25.60 & 4.29 & 3.00 & 5.53 & 0.92 & 2.49 & 4.08 & 1.35 \\
bn & 1.33 & 2.39 & 1.12 & 1.51 & 1.52 & 1.34 & 0.10 & 1.13 & 0.00 & 0.06 & 0.81 & 0.00 & 0.07 & 0.00 & 0.07 & 0.09 & 0.00 & 0.10 & 0.00 & 0.00 & 0.00 & 0.00 & 0.00 & 0.00 \\
ceb & 3.93 & 7.13 & 0.23 & 3.94 & 6.30 & 0.64 & 5.39 & 6.98 & 2.15 & 4.67 & 4.88 & 2.11 & 1.35 & 1.73 & 1.03 & 1.22 & 1.39 & 0.99 & 0.00 & 0.00 & 0.00 & 0.00 & 0.00 & 0.00\\ 
}\resultsa
\def\mystrut{\vphantom{hp}}

\begin{tikzpicture}[trim left=0cm,trim right=0cm]
    \begin{axis}[
            ybar,
            every axis plot post/.style={/pgf/number format/fixed},
            bar width=.15cm,
            width=17.5cm,
            height=4cm,
            ymajorgrids=true,
            yminorgrids=true,
            legend style={draw=none,at={(0.35,0.9)},anchor=west,font=\small},
            legend cell align={left},
            legend columns=4,
            xtick={en,fr,nl,es,ru,ja,zh,ko,vi,el,bn,ceb},
            symbolic x coords={en,fr,nl,es,ru,ja,zh,ko,vi,el,bn,ceb},
            every x tick label/.append style={font=\mystrut},
            every y tick label/.append style={font=\small\mystrut},
            tick pos=left,
            axis x line*=bottom,
            axis y line*=left,
            title={\small \hspace{.62cm} Independent: \textcolor{blue}{\small{$\blacksquare$} M-BERT} \ \textcolor{Dandelion}{\small{$\blacksquare$} XLM} \ \textcolor{ForestGreen}{\small{$\blacksquare$} XLM-R} \ \textcolor{RubineRed}{\small{$\blacksquare$} Language Specific}\\
            \small Confidence-based: \textcolor{blue}{\small{$\square$} M-BERT} \ \textcolor{Dandelion}{\small{$\square$} XLM} \ \textcolor{ForestGreen}{\small{$\square$} XLM-R} \ \textcolor{RubineRed}{\small{$\square$} Language Specific}\\
            \small \hspace{1cm}\textcolor{Gray}{$\longleftarrow$high-resource \hspace{3cm} low-resource$\longrightarrow$}
            },
            title style={align=left,yshift=-0.95cm,xshift=0cm,fill=white},
            ymin=0,ymax=18,
            ytick={0,6,12,18},
            enlarge x limits=0.05,
        ]
        \addplot [style={blue,fill=blue,bar shift=-0.4cm,}] table[x=language,y=mbertdefall]{\resultsa};
        \addplot [style={Dandelion,fill=Dandelion,bar shift=-0.15cm}] table[x=language,y=xlmdefall]{\resultsa};
        \addplot [style={ForestGreen,fill=ForestGreen,bar shift=.1cm}] table[x=language,y=xlmrdefall]{\resultsa};
        \addplot [style={RubineRed,fill=RubineRed,bar shift=.35cm}] table[x=language,y=specdefall]{\resultsa};
        \addplot [style={blue,fill=white,bar shift=-0.4cm}] table[x=language,y=mbertbestall]{\resultsa};
        \addplot [style={Dandelion,fill=white,bar shift=-0.15cm}] table[x=language,y=xlmbestall]{\resultsa};
        \addplot [style={ForestGreen,fill=white,bar shift=.1cm}] table[x=language,y=xlmrbestall]{\resultsa};
        \addplot [style={RubineRed,fill=white,bar shift=.35cm}] table[x=language,y=specbestall]{\resultsa};
        %
        %
        \addplot [style={blue,fill=blue,bar shift=-.4cm}] coordinates {(ru,1.87)};
        \addplot [style={blue,fill=blue,bar shift=-.4cm}] coordinates {(ja,0.85)};
        \addplot [style={blue,fill=blue,bar shift=-.4cm}] coordinates {(zh,2.5)};
        \addplot [style={blue,fill=blue,bar shift=-.4cm}] coordinates {(ko,4.08)};
        \addplot [style={blue,fill=blue,bar shift=-.4cm}] coordinates {(vi,8.34)};
        \addplot [style={blue,fill=blue,bar shift=-.4cm}] coordinates {(el,4.46)};
        \addplot [style={blue,fill=blue,bar shift=-.4cm}] coordinates {(bn,1.33)};
        \addplot [style={Dandelion,fill=Dandelion,bar shift=-.15cm}] coordinates {(ja,5.77)};
        \addplot [style={Dandelion,fill=Dandelion,bar shift=-.15cm}] coordinates {(el,7.1)};
        \addplot [style={ForestGreen,fill=ForestGreen,bar shift=.1cm}] coordinates {(ru,5.26)};
        \addplot [style={ForestGreen,fill=ForestGreen,bar shift=.1cm}] coordinates {(ja,2.30)};
        \addplot [style={ForestGreen,fill=ForestGreen,bar shift=.1cm}] coordinates {(zh,4.63)};
        \addplot [style={ForestGreen,fill=ForestGreen,bar shift=.1cm}] coordinates {(ko,5.11)};
        \addplot [style={ForestGreen,fill=ForestGreen,bar shift=.1cm}] coordinates {(vi,8.52)};
        \addplot [style={ForestGreen,fill=ForestGreen,bar shift=.1cm}] coordinates {(el,6.28)};
        \addplot [style={RubineRed,fill=RubineRed,bar shift=.35cm}] coordinates {(ru,6.6)};
        \addplot [style={RubineRed,fill=RubineRed,bar shift=.35cm}] coordinates {(zh,5.47)};
        \node[style={xshift=0.35cm}] at (axis cs:ceb,1) {\textcolor{RubineRed}{$\mathbf{\times}$}};
        \node[style={xshift=0.35cm}] at (axis cs:ja,1) {\textcolor{RubineRed}{$\mathbf{\times}$}};
        \node[style={xshift=0.35cm}] at (axis cs:ko,1) {\textcolor{RubineRed}{$\mathbf{\times}$}};
        \node[style={xshift=0.35cm}] at (axis cs:vi,1) {\textcolor{RubineRed}{$\mathbf{\times}$}};
        \node[style={xshift=0.35cm}] at (axis cs:bn,1) {\textcolor{RubineRed}{$\mathbf{\times}$}};
    \end{axis}
\end{tikzpicture}
\pgfplotstableread[row sep=\\,col sep=&]{
language & mbertdefall & mbertdefsingle & mbertdefmulti & mbertbestall & mbertbestsingle & mbertbestmulti & xlmdefall & xlmdefsingle & xlmdefmulti & xlmbestall & xlmbestsingle & xlmbestmulti & xlmrdefall & xlmrdefsingle & xlmrdefmulti & xlmrbestall & xlmrbestsingle & xlmrbestmulti & specdefall & specdefsingle & specdefmulti & specbestall & specbestsingle & specbestmulti \\
hu & 2.54 & 8.31 & 0.62 & 3.16 & 7.85 & 1.68 & 1.56 & 6.71 & 0.60 & 1.87 & 5.49 & 1.34 & 0.86 & 2.22 & 0.24 & 0.86 & 2.02 & 0.31 & 0.00 & 0.00 & 0.00 & 0.00 & 0.00 & 0.00 \\
he & 2.70 & 7.43 & 2.56 & 2.92 & 4.33 & 2.90 & 2.79 & 11.58 & 1.82 & 1.93 & 5.97 & 1.71 & 2.47 & 12.34 & 2.18 & 2.99 & 11.42 & 2.73 & 0.00 & 0.00 & 0.00 & 0.00 & 0.00 & 0.00 \\
tr & 2.00 & 4.50 & 1.03 & 2.08 & 4.34 & 1.19 & 1.59 & 5.53 & 0.50 & 1.85 & 4.89 & 1.06 & 3.09 & 4.04 & 2.49 & 2.95 & 3.93 & 2.43 & 3.36 & 5.88 & 2.29 & 3.13 & 5.56 & 2.15 \\
mr & 2.76 & 12.11 & 2.18 & 3.46 & 8.99 & 3.07 & 1.26 & 12.62 & 0.10 & 1.48 & 9.76 & 0.42 & 2.71 & 19.47 & 1.07 & 3.36 & 18.85 & 1.97 & 0.00 & 0.00 & 0.00 & 0.00 & 0.00 & 0.00\\ 
war & 2.29 & 4.42 & 0.42 & 2.11 & 3.73 & 0.69 & 3.29 & 5.35 & 1.83 & 1.57 & 2.17 & 1.08 & 1.15 & 1.32 & 1.08 & 1.14 & 1.29 & 1.07 & 0.00 & 0.00 & 0.00 & 0.00 & 0.00 & 0.00\\ 
tl & 5.41 & 10.12 & 0.64 & 4.62 & 7.80 & 1.25 & 4.36 & 7.35 & 1.36 & 2.25 & 3.53 & 1.11 & 2.80 & 5.05 & 1.42 & 2.33 & 4.25 & 1.28 & 0.00 & 0.00 & 0.00 & 0.00 & 0.00 & 0.00\\ 
sw & 6.24 & 10.00 & 2.25 & 6.02 & 8.42 & 3.60 & 5.90 & 8.60 & 2.18 & 4.19 & 5.90 & 2.28 & 3.66 & 5.57 & 1.91 & 2.86 & 4.34 & 1.85 & 0.00 & 0.00 & 0.00 & 0.00 & 0.00 & 0.00\\ 
pa & 1.91 & 4.35 & 1.48 & 2.56 & 3.80 & 2.30 & 0.00 & 0.00 & 0.00 & 0.00 & 0.00 & 0.00 & 0.23 & 5.75 & 0.00 & 0.58 & 5.75 & 0.36 & 0.00 & 0.00 & 0.00 & 0.00 & 0.00 & 0.00\\ 
mg & 3.36 & 4.36 & 3.27 & 3.27 & 3.40 & 3.52 & 0.00 & 0.00 & 0.00 & 0.00 & 0.00 & 0.00 & 1.94 & 3.70 & 1.61 & 1.76 & 3.49 & 1.45 & 0.00 & 0.00 & 0.00 & 0.00 & 0.00 & 0.00\\ 
yo & 3.44 & 5.15 & 3.29 & 3.21 & 3.87 & 3.06 & 0.00 & 0.00 & 0.00 & 0.05 & 0.00 & 0.05 & 0.00 & 0.00 & 0.00 & 0.01 & 0.00 & 0.01 & 0.00 & 0.00 & 0.00 & 0.00 & 0.00 & 0.00\\ 
ilo & 1.82 & 3.06 & 0.19 & 1.70 & 2.41 & 0.24 & 0.13 & 0.43 & 0.00 & 0.04 & 0.07 & 0.00 & 0.11 & 0.39 & 0.02 & 0.51 & 0.39 & 0.52 & 0.00 & 0.00 & 0.00 & 0.00 & 0.00 & 0.00 \\
}\resultsb
\def\mystrut{\vphantom{hp}}

\begin{tikzpicture}[trim left=0cm,trim right=0cm]
    \begin{axis}[
            ybar,
            every axis plot post/.style={/pgf/number format/fixed},
            bar width=.15cm,
            width=17.5cm,
            height=2.5cm,
            ymajorgrids=true,
            yminorgrids=true,
            legend style={draw=none,at={(0.35,0.9)},anchor=west,font=\small},
            legend cell align={left},
            legend columns=4,
            xtick={hu,he,tr,mr,war,tl,sw,pa,mg,yo,ilo},
            symbolic x coords={hu,he,tr,mr,war,tl,sw,pa,mg,yo,ilo},
            every x tick label/.append style={font=\mystrut},
            every y tick label/.append style={font=\small\mystrut},
            tick pos=left,
            axis x line*=bottom,
            axis y line*=left,
            ymin=0,ymax=8,
            ytick={0,4,8},
            enlarge x limits=0.05,
        ]
        \addplot [style={blue,fill=blue,bar shift=-0.4cm,}] table[x=language,y=mbertdefall]{\resultsb};
        \addplot [style={Dandelion,fill=Dandelion,bar shift=-0.15cm}] table[x=language,y=xlmdefall]{\resultsb};
        \addplot [style={ForestGreen,fill=ForestGreen,bar shift=.1cm}] table[x=language,y=xlmrdefall]{\resultsb};
        \addplot [style={RubineRed,fill=RubineRed,bar shift=.35cm}] table[x=language,y=specdefall]{\resultsb};
        \addplot [style={blue,fill=white,bar shift=-0.4cm}] table[x=language,y=mbertbestall]{\resultsb};
        \addplot [style={Dandelion,fill=white,bar shift=-0.15cm}] table[x=language,y=xlmbestall]{\resultsb};
        \addplot [style={ForestGreen,fill=white,bar shift=.1cm}] table[x=language,y=xlmrbestall]{\resultsb};
        \addplot [style={RubineRed,fill=white,bar shift=.35cm}] table[x=language,y=specbestall]{\resultsb};
        %
        %
        \addplot [style={blue,fill=blue,bar shift=-.4cm}] coordinates {(hu,2.54)};
        \addplot [style={blue,fill=blue,bar shift=-.4cm}] coordinates {(he,2.7)};
        \addplot [style={blue,fill=blue,bar shift=-.4cm}] coordinates {(tr,2)};
        \addplot [style={blue,fill=blue,bar shift=-.4cm}] coordinates {(mr,2.76)};
        \addplot [style={Dandelion,fill=Dandelion,bar shift=-.15cm}] coordinates {(tr,1.59)};
        \addplot [style={Dandelion,fill=Dandelion,bar shift=-.15cm}] coordinates {(hu,1.56)};
        \addplot [style={Dandelion,fill=Dandelion,bar shift=-.15cm}] coordinates {(mr,1.26)};
        \addplot [style={ForestGreen,fill=ForestGreen,bar shift=.1cm}] coordinates {(hu,0.86)};
        \addplot [style={ForestGreen,fill=ForestGreen,bar shift=.1cm}] coordinates {(he,2.47)};
        \addplot [style={ForestGreen,fill=ForestGreen,bar shift=.1cm}] coordinates {(mr,2.7)};
        \node[style={xshift=0.35cm}] at (axis cs:he,1) {\textcolor{RubineRed}{$\mathbf{\times}$}};
        \node[style={xshift=0.35cm}] at (axis cs:war,1) {\textcolor{RubineRed}{$\mathbf{\times}$}};
        \node[style={xshift=0.35cm}] at (axis cs:tl,1) {\textcolor{RubineRed}{$\mathbf{\times}$}};
        \node[style={xshift=0.35cm}] at (axis cs:sw,1) {\textcolor{RubineRed}{$\mathbf{\times}$}};
        \node[style={xshift=0.35cm}] at (axis cs:pa,1) {\textcolor{RubineRed}{$\mathbf{\times}$}};
        \node[style={xshift=0.35cm}] at (axis cs:mr,1) {\textcolor{RubineRed}{$\mathbf{\times}$}};
        \node[style={xshift=0.35cm}] at (axis cs:yo,1) {\textcolor{RubineRed}{$\mathbf{\times}$}};
        \node[style={xshift=0.35cm}] at (axis cs:mg,1) {\textcolor{RubineRed}{$\mathbf{\times}$}};
        \node[style={xshift=0.35cm}] at (axis cs:ilo,1) {\textcolor{RubineRed}{$\mathbf{\times}$}};
        \node[style={xshift=0.1cm}] at (axis cs:yo,1) {\textcolor{ForestGreen}{$\mathbf{\times}$}};
        \node[style={xshift=-0.15cm}] at (axis cs:yo,1) {\textcolor{Dandelion}{$\mathbf{\times}$}};
        \node[style={xshift=-0.15cm}] at (axis cs:mg,1) {\textcolor{Dandelion}{$\mathbf{\times}$}};
        \node[style={xshift=-0.15cm}] at (axis cs:pa,1) {\textcolor{Dandelion}{$\mathbf{\times}$}};
    \end{axis}
\end{tikzpicture}
\caption{Accuracy on different languages using different LMs (\%). 
Independent prediction (solid bars) outperforms confidence-based prediction (no-fill bars) on high-resource languages but not on low-resource languages. Different models are color-coded, with missing/unsupported models marked with $\mathbf{\times}$. Languages are ranked by the total number of facts in our benchmark. Details in Appendix \autoref{tab:multilang}.}
\label{fig:multilang}
\end{figure*}
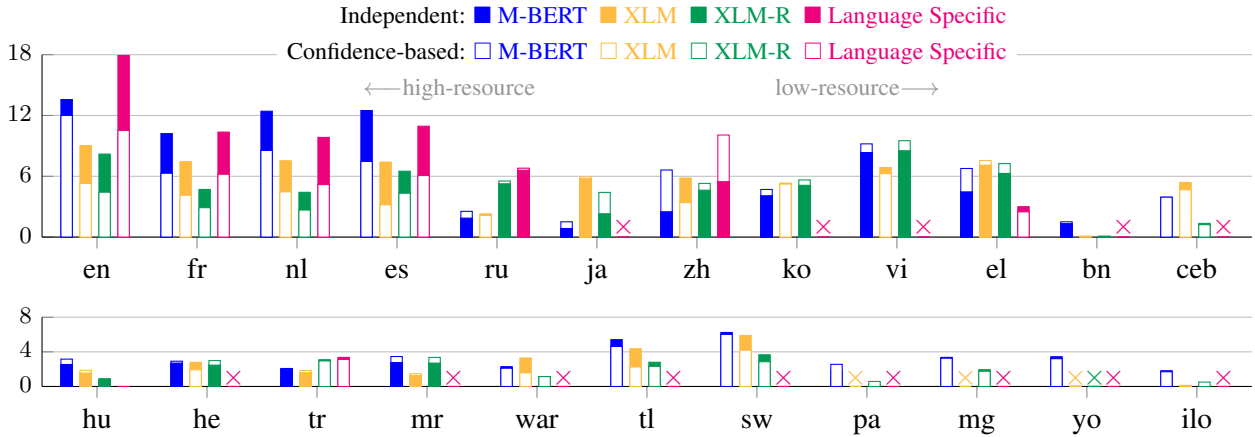

We run both the independent and confidence-based decoding methods with 3 M-LMs, and when available 8 monolingual LMs, across 23 languages,\footnote{Check \url{https://x-factr.github.io} for latest results.} with results shown in \autoref{fig:multilang}.
Overall, even in the most favorable settings, the performance of state-of-that-art M-LMs at retrieving factual knowledge in the \methodabbr benchmark is relatively low, achieving less than 15\% on high-resource languages (e.g., English and Spanish) and less than 5\% for some low-resource languages (e.g., Marathi and Yoruba).
This may initially come as a surprise, given the favorable performance reported in previous papers \cite{petroni-etal-2019-language,jiang-2019-lpaqa}, which achieved accuracies over 30\% on English.
We justify this discrepancy in our following analysis. We note that, although we provide baseline results in almost all languages, we perform our extensive analysis on a representative subset, consisting of 13 languages.

\paragraph{Performance on Different Languages.}
Performance on high-resource languages is usually better than performance on middle- or low-resource languages regardless of the (M-)LMs.
This is probably due to high-resource languages having more data in the pre-training stage.
It is also possible that even if the fact of low-resource languages is written in the available data for these languages, it is not appropriately memorized due to lack of model capacity or forgetting \cite{kirkpatrick2017overcoming}.
It is worth noting that the best results are in Indo-European languages which not only have the most data, but also share the same (Latin) script which could further facilitate cross-lingual learning.

\paragraph{Performance of Different LMs.}
Comparing the performance of different M-LMs, we found that M-BERT outperforms XLM and XLM-R on high-resource languages, while on low-resource languages performance is similar.
This is contradictory to the conclusion on other cross-lingual tasks, such as natural language inference and syntactic prediction, as reported in \citet{hu-2020-xtreme}.
Our conjecture is that because factual knowledge probing requires retrieving the identity and relations of individual entities, it is more fine-grained than more coarse-grained understanding of syntactic and semantic classes that are required to solve other tasks.
We posit that pre-training methods that show superior performance on inference and syntactic prediction tasks (i.e., XLM-R) might achieve good syntactic/semantic abstraction at the cost of making less concrete lexical distinctions.

Comparing M-BERT with language-specific LMs, we find M-BERT outperforms the monolingual BERT on Dutch, Spanish, and Greek, while underperforming on English, Russian, Chinese, and Turkish.
Since most of the LMs follow the architecture and pre-training settings of BERT \cite{devlin-etal-2019-bert} or RoBERTa \cite{liu-2019-roberta}, we hypothesize that training corpus is the major contributor to the final performance, and summarize those corpora in \autoref{tab:shortcut} in the Appendix.
Another potential explanation is that model capacity limitations preclude M-BERT from effectively memorizing entity names/relations in all of the languages.

\begin{figure}[t]
\pgfplotstableread[row sep=\\,col sep=&]{
lang & nooracle & withoracle & single \\
en & 12 & 19.89 & 25.45 \\
fr & 6.3 & 13.32 & 20.55 \\
nl & 8.55 & 14.8 & 26.08 \\
es & 7.47 & 16.76 & 25.1 \\
ru & 2.54 & 4.82 & 5.42 \\
zh & 6.62 & 12.52 & 10.13 \\
he & 2.92 & 5.76 & 9.61 \\
tr & 2.08 & 2.59 & 5.08 \\
ko & 4.7 & 8.17 & 23.23 \\
vi & 9.2 & 15.55 & 18.49 \\
el & 6.77 & 12.04 & 23.97 \\
mr & 3.46 & 5.52 & 13.5 \\
yo & 3.21 & 4.04 & 5.15\\
}\oracledata
\def\mystrut{\vphantom{hp}}

\hspace{.3cm}
\begin{tikzpicture}[trim left=0cm,trim right=0cm]
    \begin{axis}[
            ybar,
            every axis plot post/.style={/pgf/number format/fixed},
            bar width=.07cm,
            width=8.8cm,
            height=3.5cm,
            ymajorgrids=true,
            yminorgrids=true,
            legend style={draw=none,at={(0.35,0.9)},anchor=west,font=\small},
            legend cell align={left},
            legend columns=3,
            xtick={en,fr,nl,es,ru,zh,he,tr,ko,vi,el,mr,yo},
            symbolic x coords={en,fr,nl,es,ru,zh,he,tr,ko,vi,el,mr,yo},
            every x tick label/.append style={font=\mystrut},
            every y tick label/.append style={font=\tiny\mystrut},
            tick pos=left,
            axis x line*=bottom,
            axis y line*=left,
            title={\small \textcolor{blue}{\small{$\blacksquare$}} w/o oracle \\[-.1cm] \small \textcolor{Dandelion}{\small{$\blacksquare$}} with oracle \\[-.1cm] \small \textcolor{ForestGreen}{\small{$\blacksquare$}} single-token},
            title style={align=left,yshift=-1.3cm,xshift=0cm,fill=white},
            ymin=0,ymax=26,
            ytick={0,10,20},
            enlarge x limits=0.05,
        ]
        \addplot [style={blue,fill=blue,}] table[x=lang,y=nooracle]{\oracledata};
        \addplot [style={Dandelion,fill=Dandelion,}] table[x=lang,y=withoracle]{\oracledata};
        \addplot [style={ForestGreen,fill=ForestGreen,}] table[x=lang,y=single]{\oracledata};
    \end{axis}
\end{tikzpicture}
\caption{Accuracy of the confidence-based decoding algorithm on different languages using M-BERT w/ and w/o oracle length (\%).}
\label{fig:oracle_len}
\end{figure}
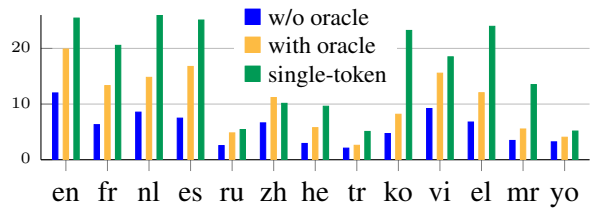

\paragraph{Single-token vs Multi-token.}
Since we choose among $M$ candidate predictions with different numbers of mask tokens based on confidence, it is possible that the prediction with the correct number of mask tokens has lower confidence than the other predictions.
To investigate the errors introduced by this step, we conduct an ablation experiment that assumes we know the ground-truth number of mask tokens.
As shown in \autoref{fig:oracle_len}, performance improves significantly by 75\% on average across all languages using the oracle mask number, indicating that pre-trained LMs have difficulties in choosing the correct number of mask tokens.
The performance on single-token facts (i.e., the setting of previous works that only predicts a single token) is even higher, demonstrating the difficulty of multi-token prediction.\footnote{The 31.1\% accuracy of BERT in \citet{petroni-etal-2019-language} is over a different set of facts in English, constrained to be in the intersection of vocabularies of several LMs. We have no such constraint, which may explain the slightly lower 25.5\% accuracy on the English single-token performance in \autoref{fig:oracle_len}.}

\begin{table*}
\vspace{-5mm}
\scriptsize
\centering
\begin{tabular}{@{}l@{\tinycol}l@{\tinycol}l@{\tinycol}l@{\tinycol}r@{\tinycol}r@{\tinycol}r@{}}
\toprule
\textbf{Type} & \textbf{Prompt} & \textbf{Prediction} & \textbf{Gold} & \multicolumn{1}{c}{\textbf{en}} & \multicolumn{1}{c}{\textbf{es}} & \multicolumn{1}{c}{\textbf{el}} \\
\midrule
\emph{Correct} & Macintosh 128K is produced by \_. & Apple & Apple & \emph{19.89} & \emph{16.68} & \emph{12.02} \\
\midrule
Repeating subjects & Malin Reuterwall plays with \_. & the Reuterwall team & Sweden's Womens Football & 22.21 & 24.62 & 25.06 \\
Wrong entities & Austria maintains diplomatic relations with \_. & the United States & Italy, Russia, ... & 16.66 & 29.07 & 18.74 \\
Non-informativeness & Switzerland is named after \_. & him & Canton of Schwyz & 18.24 & 9.81 & 26.78 \\
Type errors & Nin9 2 5ive was written in \_. & the 1880s & Cantonese & 7.93 & 6.11 & 0.00 \\
Related concepts & Christof Lauer used to work in \_. & Germany & Melsungen & 7.14 & 1.67 & 1.91 \\
Unk & Randy Newman plays \_. & D.D & piano & 5.55 & 8.33 & 11.67 \\
False Negative & Switzerland maintains diplomatic relations with \_. & the Federal Republic of Germany & Germany & 2.38 & 3.52 & 3.06 \\
Inflection & - & - & - & 0.00 & 0.19 & 0.77\\
\bottomrule
\end{tabular}
\caption{Error cases of M-BERT in English and ratio of different error types in English, Spanish, and Greek (\%). Error cases in Spanish and Greek can be found in \autoref{tab:err_more} in the Appendix.}
\label{tab:err}
\vspace{-0.5cm}
\end{table*}

\paragraph{Error Analysis.}
Even with access to an oracle for the number of target tokens, though, the performance is still lower than 20\%.
To understand the types of errors made by the LMs, we sample over 400 error cases in English, Spanish, and Greek, and classify them.
The error type distributions along with English examples are outlined in \autoref{tab:err}.

The most prominent error type, about one-fourth of mistakes for all LMs, was \textbf{repeating subjects}, whereby the prediction repeats either the full or partial subject.
Predicting the \textbf{wrong entities} is also fairly common, especially in Spanish (29\%). Interestingly, we find that wrong predictions are often a language-specific ``common" entity such as `{\textgreekfont Αθήνα}' (Athens, the capital of Greece) in Greek location prompts, while the Spanish model insisted most musicians play `flauta' (flute).
Another error type, particularly common in Greek (27\%), is producing \textbf{non-informative} output, where the predictions are function words that could never be an entity.
\textbf{Type errors} when the semantic type of the prediction is different than expected (e.g. predicting dates instead of locations) are fairly common (English: 8\%, Spanish 6\%), as are \textbf{related concepts} predictions (English: 7\%), where the model predicts relevant, possibly factually correct entities (e.g. predicting a country or a state instead of a city).
Worryingly, in a fair amount of cases (English: 5\%, Spanish: 8\%, Greek: 11\%) the models output non-existent words (\textbf{unk}).
Errors of the last 4 types could potentially be avoided by limiting the allowed outputs of the model to specific entity classes; we leave this for future work.
Last, we identified around 3\% of \textbf{false negatives}, where the prediction is actually correct but is not part of our aliases list and less than 1\% of \textbf{inflection} errors where the prediction is the correct entity but improperly inflected.

\paragraph{Performance of Different Decoding Methods.}\label{sec:exp_dec}
Overall, the confidence-based decoding method improves the accuracy in middle- and low- resource languages, while it hurts the performance on high-resource languages.
To better understand the effect of different components on the final performance, we conduct a comprehensive comparison on English and Chinese.
We compare the three initial prediction methods and the three refinement options (including not performing refinement), for a total of nine decoding methods (\autoref{sec:dec}).
We further apply additional improvements (\autoref{sec:dec_add}) on the confidence-based decoding method.

\begin{table}[t]
\small
\centering
\begin{tabular}{@{}l@{\tinycol}l@{\tinycol}r@{\tinycol}r@{\tinycol}r|r@{\tinycol}r@{\tinycol}r@{}}
\toprule
 & & \multicolumn{3}{c|}{\textbf{English}} & \multicolumn{3}{c}{\textbf{Chinese}} \\
\textbf{Init.} & \textbf{Refine} & \multicolumn{1}{c}{\textbf{All}} & \textbf{Single} & \textbf{Multi} &  \multicolumn{1}{c}{\textbf{All}} & \textbf{Single} & \textbf{Multi} \\
\midrule
Indep. & - & 13.57 & \textbf{22.40} & 5.57 & 2.50 & \textbf{9.61} & 2.22 \\
 & Order & \textbf{13.91} & 21.71 & 6.71 & 4.26 & 8.80 & 4.01 \\
 & Conf. & 13.38 & 21.49 & 5.82 & 4.04 & 9.33 & 3.80 \\
\midrule
Order & - & 13.54 & 20.37 & 6.60 & 5.06 & 8.57 & 4.85 \\
 & Order & 13.30 & 19.75 & 6.57 & 5.79 & 8.29 & 5.61 \\
 & Conf. & 13.36 & 19.86 & 6.56 & 5.68 & 8.29 & 5.50 \\
\midrule
Conf. & - & 13.64 & 19.53 & 7.38 & 6.55 & 5.34 & 6.41 \\
 & Order & 13.73 & 19.48 & 7.57 & \textbf{6.79} & 4.63 & \textbf{6.67} \\
 & Conf. & 13.72 & 19.44 & 7.48 & 6.62 & 5.21 & 6.40 \\
\midrule
\midrule
\multicolumn{2}{r}{+Len. norm} & 8.60 & 9.43 & 6.18 & 3.96 & 2.27 & 3.93 \\
\multicolumn{2}{r}{+Re-comp.} & 12.00 & 12.91 & 10.08 & 5.89 & 2.71 & 5.84 \\
\multicolumn{2}{r}{+Beam} & 10.84 & 9.29 & \textbf{11.06} & 6.34 & 2.38 & 6.30 \\
\bottomrule
\end{tabular}
\caption{Accuracy of different decoding methods using M-BERT on English and Chinese (\%).}
\label{tab:decode}
\end{table}

By comparing the performance in \autoref{tab:decode}, we first see advanced decoding methods improve performance on multi-token objects, but hurt performance on single-token ones.
The best-performing decoding method on English improves the multi-token accuracy from 5.57\% to 11.06\%, indicating that advanced decoding methods have a better chance to elicit multi-token facts from M-BERT.
Some examples are shown in \autoref{tab:decode_en_case} in the Appendix.
The lower performance on single-token objects is probably caused by the fact that advanced decoding methods discover multi-token predictions that have higher confidence than single-token ones (\autoref{eq:conf}).
For example, the single-token prediction for ``Enrique Iglesias used to communicate in \_.'' is ``Spanish'', while the best decoding method outputs ``his own words'' with higher confidence.
Second, initial prediction methods have a greater effect on the final performance than refinement methods.
We hypothesize that this is because the greedy decoding process heavily depends on previous predictions, and refinement cannot recover from unsatisfactory initial predictions.
Third, length normalization was not found useful in either case.

There are also observations not consistent across the two languages.
First, since Chinese has a larger portion of multi-token objects than English (as shown in \autoref{tab:data_stat}), the overall performance on Chinese increases while it decreases on English, which is consistent with the observation in \autoref{fig:multilang}.
Second, confidence re-computation and beam search are not as effective on Chinese, which we conjecture is because that the distribution over English sentences exhibits more multimodality than the distribution over Chinese sentences due to more training data.

\section{Improving Multilingual LM Retrieval}
\label{sec:code_switch}
\begin{figure}[t]
\centering
\includegraphics[width=1.0\columnwidth, clip, keepaspectratio]{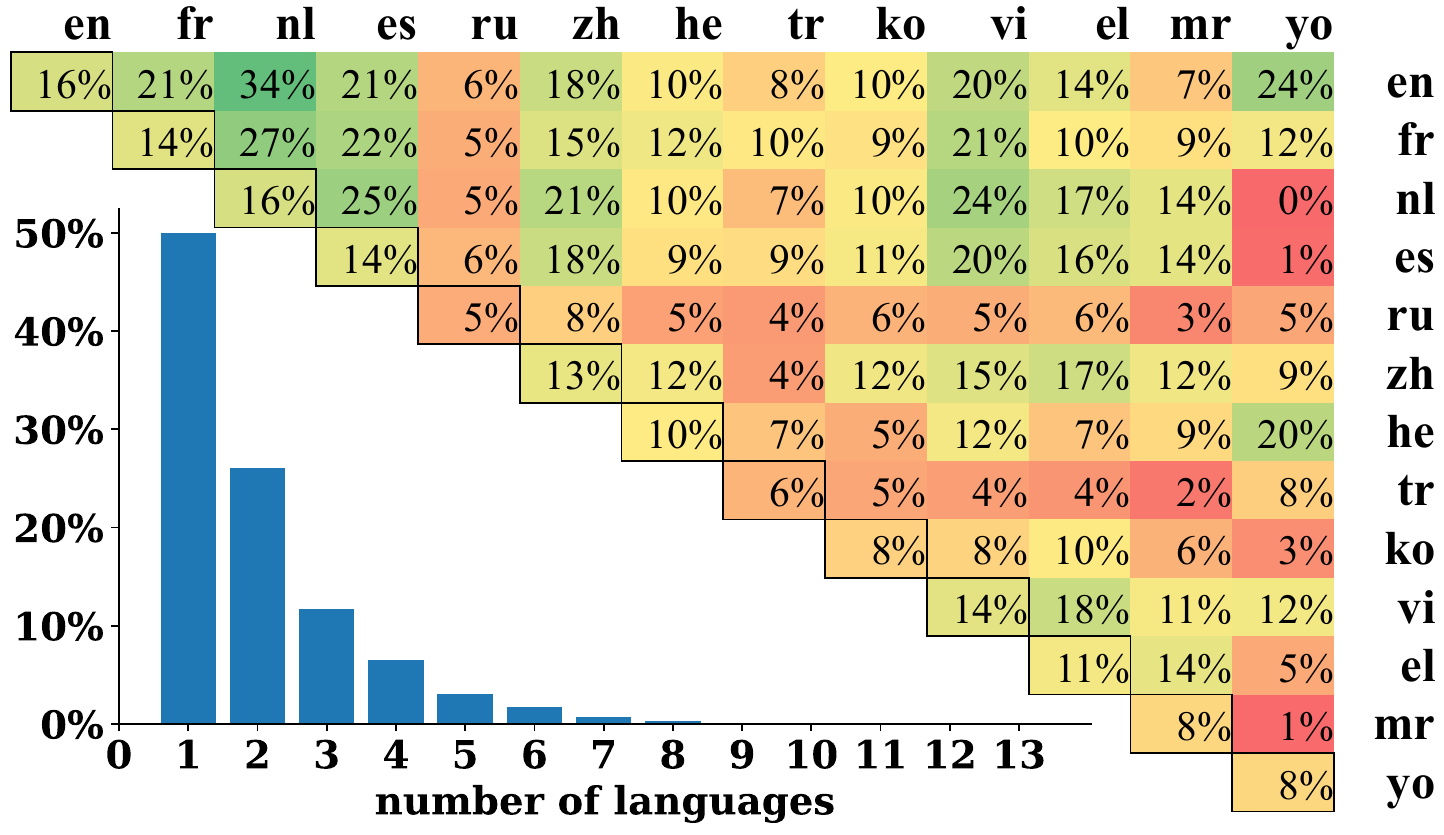}
\caption{Bottom-left: the ratio of facts with respect to the number of languages in which the facts could be successfully retrieved. Top-right: overlap ratio of correct predictions between two languages. The values on the diagonal are the average overlap ratio of the corresponding language with the other languages.}
\label{fig:overlap}
\end{figure}
As the performance of M-LMs is relatively low, especially on low-resource languages, an obvious endeavor is to refine the model to improve fact retrieval performance in various languages.
We analyze how similarly M-BERT performs on queries in different languages.
We collect correctly predicted facts across all languages, and count in how many languages each fact was retrieved correctly.
As shown in the bottom-left histogram of \autoref{fig:overlap}, half of the correctly predicted facts were correct in a single language, indicating little overlap across languages \cite{lin2018social}.
Only 3\% of facts were correct in more than 5 languages, and objects in those facts are usually sub-strings of subjects, making them easy to retrieve regardless of the language.
This observation is also confirmed by the overlap between pairs of languages in the top-right chart of \autoref{fig:overlap}; even the most similar languages (i.e., English and Dutch) only have 34\% of correct predictions in common.

We find that facts retrievable only in a single language tend to be knowledge that is mainly mentioned in a certain language.
For example, M-BERT mistakenly predicts ``QQ'' in the English sentence ``Tencent QQ is developed by \_.'', while the prediction ``\begin{CJK*}{UTF8}{gbsn}腾讯\end{CJK*}'' (Tencent) in the corresponding Chinese sentence ``\begin{CJK*}{UTF8}{gbsn}腾讯QQ是由\_开发的。\end{CJK*}'' is correct.
This is probably because Tencent, a Chinese company, is more frequently mentioned in the Chinese training corpus.

\subsection{Methods}
Inspired by these observations, we propose to use \emph{code-switching} to create data to fine-tune pre-trained LMs, replacing entity mentions in one language (e.g., English/Greek) with their counterparts in another language (e.g., Greek/English).
Through this bi-directional code-switching, entity mentions serve as pivots, enabling knowledge that was originally learned in one language to be shared with others.
Given a pair of languages, we first identify Wikipedia sentences that mention entities from our benchmark using SLING \cite{ringgaard-2017-sling}. The M-LM is then finetuned on these sentences. Following \citet{wu-2019-crosslingual}, with 30\% of probability we switch all the entity mentions (can be one or multiple) from the original language to their counterparts in the other language, ending up with sentences like ``{\textgreekfont Oμπάμα} later reflected on his years ...", where we substituted ``Obama" with a Greek mention of the entity, and vice-versa for Greek-to-English.
70\% of the sentences remain the same.
If there are multiple mention texts for an entity, we sample proportionally to their frequencies, which we found in our preliminary experiments performed better than using a fixed translation.
We fine-tune M-BERT using the masked LM objective on this data, with 15\% of non-mention words and 50\% of mention words masked out.\footnote{The larger ratio on entities encourages the model to focus on predicting entities, as in the downstream task.}

\subsection{Experimental Results}
We choose three languages with different data availability, namely French, Russian, and Greek, and pair them with English, producing 560k, 396k, and 129k code-switched sentences respectively.
We compare M-BERT after code-switched fine-tuning (denoted as \textbf{cs}) with both the original M-BERT and with fine-tuning only on raw text (\textbf{raw}).
We vary the evaluation settings to illustrate the effect of code-switching: on top of matching predictions to ground truth aliases in the prompt language (\textbf{single-eval)}, we evaluate with targets in both languages (\textbf{double-eval}; English and prompt).

\begin{table}[t]
\small
\begin{center}
\begin{tabular}{@{}l@{\tinycol}l@{\tinycol}r@{\tinycol}r@{\tinycol}r@{\tinycol}r@{\tinycol}r@{\tinycol}r@{}}
\toprule
 & & \multicolumn{3}{c}{\textbf{Single-eval}} & \multicolumn{3}{c}{\textbf{Double-eval}} \\
\textbf{Lang.} & \textbf{Method} & \multicolumn{1}{c}{\textbf{All}} & \textbf{Single} & \textbf{Multi} &  \multicolumn{1}{c}{\textbf{All}} & \textbf{Single} & \textbf{Multi} \\
\midrule
\multirow{3}{*}{French} & M-BERT & 10.21 & 19.07 & 3.92 & 10.67 & 19.24 & 4.55 \\
 & \quad+raw & \textbf{15.06} & \textbf{26.81} & \textbf{7.40} & 15.69 & 26.92 & 8.27 \\
 & \quad+cs & 13.15 & 24.37 & 6.34 & \textbf{16.90} & \textbf{26.98} & \textbf{10.29} \\
\midrule
\multirow{3}{*}{Russian} & M-BERT & 1.87 & 4.58 & 0.96 & 3.04 & 7.72 & 2.28 \\
 & \quad+raw & \textbf{7.92} & \textbf{24.37} & \textbf{3.59} & 8.77 & \textbf{26.28} & 4.57 \\
 & \quad+cs & 7.64 & 22.41 & 3.55 & \textbf{11.69} & 25.31 & \textbf{7.85} \\
\midrule
\multirow{3}{*}{Greek} & M-BERT & 4.49 & 20.75 & 2.19 & 4.97 & 20.87 & 2.83 \\
 & \quad+raw & \textbf{11.49} & \textbf{35.27} & \textbf{7.65} & 12.65 & \textbf{35.27} & 9.27 \\
 & \quad+cs & 9.30 & 26.31 & 5.73 & \textbf{18.41} & 30.93 & \textbf{15.30} \\
\bottomrule
\end{tabular}
\end{center}
\caption{Accuracy of M-BERT after fine-tuning on raw and code-switched text (\%).}
\label{tab:cs}
\end{table}

As shown in \autoref{tab:cs}, continued fine-tuning on raw text outperforms the original M-BERT, likely due to our fine-tuning on a subset of sentences with mentions of entities from our benchmark.
Results on code-switched text are slightly worse when only matching entities in the original target language, but significantly better if we allow matching in both the original language and English.
This indicates that code-switched fine-tuning allows M-BERT to retrieve facts, albeit in English rather than in the prompt language.
Encouragingly, the increase is larger for low-resource (Greek) and typologically distant-to-English (Russian) languages.
For example, the prediction for the Greek prompt ``{\textgreekfont η Θεωρία κατηγοριών είναι μέρος των \_.}'' (``Category theory is part
of \_.'') is ``mathematics'' (in English!), while the prediction without code-switching is the non-informative ``{\textgreekfont οποίων}'' (``which'').
Considering that we have more raw than code-switched sentences in the dataset, this seems to indicate that English entities are easier to predict than their prompt-language counterparts, which might be because facts expressed in English are better learned in the pre-trained model due to training data abundance.

\section{Related Work}
\label{sec:related}
\paragraph{Factual Knowledge Retrieval from LMs}
Several works have focused on probing factual knowledge solely from pre-trained LMs without access to external knowledge.
They do so by either using prompts and letting the LM fill in the blanks, which assumes that the LM is a static knowledge source \cite{petroni-etal-2019-language,jiang-2019-lpaqa,poerner-2019-ebert,bouraoui-2020-relbert}, or fine-tuning the LM on a set of question-answer pairs to directly generate answers, which dynamically adapts the LM to this particular task \cite{roberts-2020-t5pack}.
Impressive results demonstrated by these works indicate that large-scale LMs contain a significant amount of knowledge, in some cases even outperforming competitive question answering systems relying on external resources \cite{roberts-2020-t5pack}.
\citet{petroni-2020-contextlm} further shows that LMs can generate even more factual knowledge when augmented with retrieved sentences.
Our work builds on these works by expanding to multilingual and multi-token evaluation, and also demonstrates the significant challenges posed by this setting.

\paragraph{Multilingual Benchmarks}
Many multilingual benchmarks have been created to evaluate the performance of multilingual systems on different natural language processing tasks, including question answering \cite{artetxe-2019-cross,lewis-2019-mlqa,clark-2020-tydiqa}, natural language understanding \cite{conneau-etal-2018-xnli,yang-etal-2019-paws,zweigenbaum-2018-overview,artetxe-2019-massively}, syntactic prediction \cite{nivre-2018-universal,pan-etal-2017-cross}, and comprehensive benchmarks covering multiple tasks \cite{hu-2020-xtreme,liang-2020-xglue}.
We focus on multilingual factual knowledge retrieval from LMs, which to our knowledge has not been covered by any previous work.

\section{Conclusion}
We examine the intersection of multilinguality and the factual knowledge included in LMs by creating a multilingual and multi-token benchmark \methodabbr, and performing experiments comparing and contrasting across languages and LMs.
The results demonstrate the difficulty of this task, and that knowledge contained in LMs varies across languages.
Future directions include other pre-training or fine-tuning methods to improve retrieval performance and methods that encourage the LM to predict entities of the right types.

\section*{Acknowledgements}
This work was supported by a gift from Bosch Research.
The authors are thankful to the reviewers for the thorough and insightful comments.
They are also particularly grateful for everyone who helped create, check, or evaluate the templates and the outputs of our models: Aman Madaan, Aditi Chaudhary, Paul Michel, Sergio Franco, Maria Ryskina, Chan Young Park, Hiroaki Hayashi, Toan Nguyen, David Ifeoluwa Adelani, Bonaventure Dossou, Emre Yolcu, Happy Buzaaba, and Fahim Faisal.

\bibliography{emnlp2020}
\bibliographystyle{acl_natbib}

\clearpage
\newpage
\appendix

\section{Benchmark Details}

\begin{figure*}
\centering
\includegraphics[width=1.0\textwidth, clip, keepaspectratio]{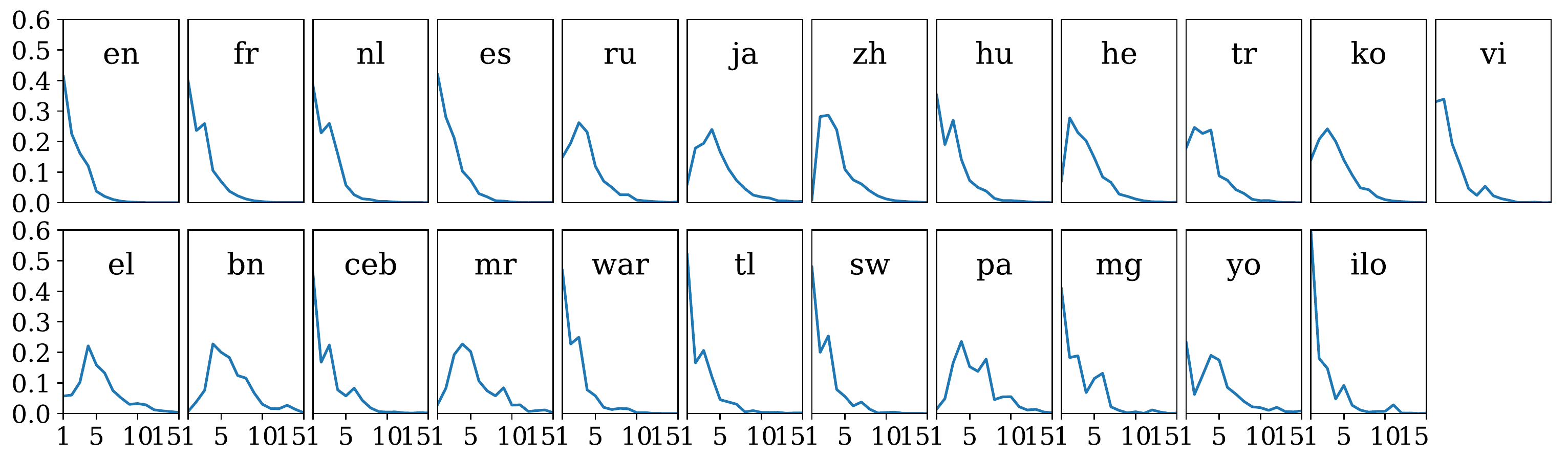}
\caption{Ratio of facts with respect to the number of tokens of the object in different languages.}
\label{fig:token}
\end{figure*}

\autoref{tab:detail_data_stat} shows the detailed number of facts in each language in our \methodabbr benchmark.
\autoref{fig:token} demonstrates the ratio of facts with respect to the number of tokens of the object in different languages, where high-resource languages (e.g., English, French, Dutch, and Spanish) have more portion of single-token facts than low-resource languages.

\begin{table*}[t]
\small
\begin{center}
\begin{tabular}{l@{\smallcol}r@{\smallcol}r@{\smallcol}r@{\smallcol}r@{\smallcol}r@{\smallcol}r@{\smallcol}r@{\smallcol}r@{\smallcol}r@{\smallcol}r@{\smallcol}r@{\smallcol}r@{\smallcol}r}
\toprule
& en & fr & nl & es & ru & ja & zh & hu & he & tr & ko & vi \\ 
 \#facts & 45684 & 40240 & 38291 & 37065 & 26265 & 25144 & 23142 & 20438 & 17050 & 16104 & 16098 & 13642 \\ 
 \#single-word facts & 18903 & 13886 & 12812 & 13463 & 3391 & 1312 & 210 & 6241 & 1057 & 2506 & 1964 & 3909 \\ 
 \#multi-word facts & 26781 & 26354 & 25479 & 23602 & 22874 & 23832 & 22932 & 14197 & 15993 & 13598 & 14134 & 9733 \\
\midrule
& el & bn & ceb & mr & war & tl & sw & pa & mg & yo & ilo \\
\midrule
\#facts & 13034 & 9383 & 8160 & 7877 & 7342 & 7116 & 6834 & 5455 & 4945 & 4609 & 4053 \\
\#single-word facts & 742 & 53 & 3257 & 199 & 2981 & 3208 & 2840 & 67 & 1748 & 930 & 2099 \\
\#multi-word facts & 12292 & 9330 & 4903 & 7678 & 4361 & 3908 & 3994 & 5388 & 3197 & 3679 & 1954 \\
\bottomrule
\end{tabular}
\end{center}
\caption{Detailed \methodabbr Benchmark statistics. Languages are ranked by the total number of facts.}
\label{tab:detail_data_stat}
\end{table*}

\section{Benchmark Prompt Quality}
\label{app:prompt}
The prompts generated in different languages may not be perfectly natural.
This could be due to awkwardness of attempting to express relational phrases that were originally devised for English in languages where the semantic distinctions of the underlying words may differ, or due to our errors in our automated approach to grammatical attribute inference and subsequent inflection.
To this end, we evaluated our prompts on a sample of languages, providing native speakers with~10 sentences per prompt with the missing slots filled by our inflection models. Our approach produces sentences that are annotated as correct 97.9\% of the cases in Spanish, 90.5\% in Yoruba, 86.7\% in Greek, 82.3\% in Marathi, and 81.9\% in Russian. 

\begin{table}[t]
\scriptsize
\centering
\begin{tabular}{@{}c@{\tinycol}c@{\tinycol}c@{\tinycol}c@{\tinycol}c@{\tinycol}c@{\tinycol}c@{}}
\toprule
    & & \multicolumn{5}{c}{\% Errors}\\
    Language & \% Correct & Inflection & Gender & Number & Awkward & Wrong Sense \\
\midrule
    Greek & 86.7 & 5.4 & 7.4 & 0.5 & 5.0 & 5.0 \\
    Spanish & 97.9 & -- & 1.6 & 0.8 & 1.9 & 0  \\
    Marathi & 82.3 & 15.1 & -- & 0.2 & 0 & 4\\
    Russian & 81.9 & 16.1* & -- & -- & 18.1* & 6.7 \\
    Yoruba & 90.5 & -- & -- & -- & 4.1 & 0 \\
\bottomrule
\end{tabular}
\caption{Error analysis on the prompts after instantiating with actual examples. We note that the error categories are not mutually exclusive. *: The Russian inflection percentage includes gender and number errors, unlike the other languages; the Russian annotator also marked all erroneous sentences as ``awkward", skewing the results.}
\label{tab:app_prompts}
\end{table}

We present an analysis of the annotations on the erroneous prompts in Table~\ref{tab:app_prompts}. The error types differ drastically across languages. Russian and Marathi have comparatively large percentages of inflection-related errors, but for different reasons: the prediction of non-human entity grammatical gender in Russian is difficult and this results in mistakes in the inflection. In Marathi, this issue is also exacerbated by the inflection model, which is of slightly lower quality due to the scarcity of training data availability.

Despite these two outliers, we consider the rest of our prompts to be of high quality. Even if small inflection or grammatical gender assignment mistakes occur (e.g. in Greek) this should not render the prompt unintelligible to native speakers -- the burden is on the model to be robust to such slight variations, just as humans are.
We point out that the prompts can be awkward or incorrect for some senses captured by the relation, an issue unrelated to our gender heuristics or automatic inflection. This issue, though, is also present in the LAMA English prompts~\cite{petroni-etal-2019-language,jiang-2019-lpaqa} and is the result of the original Wikidata annotation.

\section{Multi-Token Decoding}\label{ap:decoding}

We outline here the exact concrete formulation of our multi-token decoding algorithms.
Given a sentence with multiple mask tokens, e.g., \autoref{eq:prompt}, we can either generate outputs in parallel independently or one at a time conditioned on the previously generated tokens.
These methods are similar to the prediction problems that BERT \cite{devlin-etal-2019-bert} and XLNet \cite{yang-2019-xlnet} perform in their pre-training stages respectively.
We define $\bm{c} \in \mathbb{R}^n$ as the confidence of each prediction, with details varying by prediction method.

\subsection{Initial Prediction and Refinement}
\paragraph{Independent}
For independent initial prediction, the mask tokens are all predicted in parallel:
\begin{align*}
\hat{y}_k =& \argmax_{y_k} p(y_k|\bm{s}_{i:j}), c_k = p(\hat{y}_k|\bm{s}_{i:j}),\\
&\forall k \in \{i, ..., j\}.
\end{align*}
We also consider two autoregressive methods for initial prediction or refinement.

\paragraph{Order-based} Mask tokens are predicted from left to right, conditioned on previously generated tokens in each step:
\begin{equation*}
\small
\hat{y}_i = \argmax_{y_i} p(y_i|\bm{s}_{i:j}), c_i = p(\hat{y}_i|\bm{s}_{i:j}).
\end{equation*}
In the refinement stage, we modify the predicted tokens from left to right by replacing the token with a \mask and re-predicting it:
\begin{equation*}
\hat{y}_i = \argmax_{y_i} p(y_i|\hat{\bm{s}}_{i:j}\setminus i), c_i = p(\hat{y}_i|\hat{\bm{s}}_{i:j}\setminus i),
\end{equation*}
where $\bm{s}\setminus i$ means that the $i$-th token in $\bm{s}$ is replaced with \mask.
Convergence is reached when there are no changes in a left-to-right scan.

\paragraph{Confidence-based} Among all the predictions for masked positions, we choose the one with the highest confidence (i.e., the highest probability), so the actual order of predictions can be arbitrary, as shown in \autoref{fig:decode}:
\begin{equation*}
\hat{y}_k = \argmax_{i \leq k\leq j, y_k} p(y_k|\bm{s}_{i:j}), c_k = p(\hat{y}_k|\bm{s}_{i:j}).
\end{equation*}
In the refinement stage, we choose from all predicted tokens the one with the lowest confidence (i.e., the lowest probability) and re-predict it \cite{ghazvininejad-etal-2019-mask}:
\begin{align*}
    \hat{y}_k =& \argmax_{y_k} p(y_k|\hat{\bm{s}}_{i:j}\setminus k), c_k = p(\hat{y}_k|\hat{\bm{s}}_{i:j}\setminus k),\\
    &k = \argmin_{i\leq k\leq j} c_k. 
\end{align*}
Convergence is reached when the re-predicted token is the same as the original token.
\fi

\subsection{Additional Decoding Components}\label{ap:dec_add}
\paragraph{Length Normalization}
Since the sum used in \autoref{eq:conf} might favor short predictions, we consider normalizing it by the number of the mask tokens:
\begin{equation*}
\small
v(j-i+1) = \frac{1}{j-i+1}\sum_{k=i}^{j}{\log c_k},
\end{equation*}

\paragraph{Confidence Re-computation}
Note that the confidence of each predicted token $\bm{c}$ in previous equations is the probability when the token is predicted.
However, the probability will become stale once the surrounding tokens change because of the bidirectional conditional distributions, and this is also noted in \citep{ghazvininejad-etal-2019-mask}.
To make the confidence up-to-date, given the prompt in \autoref{eq:prompt}, when a new token is predicted (in the initial stage) or a token is modified (in the refinement stage), we re-compute $c_i$ to $c_j$.
This makes the time complexity quadratic to the number of mask tokens, because every time we make a modification, we have to re-compute the confidence values of all predictions.
As a result, the final confidence becomes:
\begin{equation*}
c_k = p(\hat{y}_k|\hat{\bm{s}}_{i:j}\setminus k),
\end{equation*}
where $\hat{\bm{s}}_{i:j} = x_1, ..., \hat{y}_i, ..., \hat{y}_j, ..., x_n$ contains the final predictions.

\paragraph{Beam Search}
All of the previous methods use the most plausible prediction at each masked position.
We also consider performing beam search that keeps track of the most plausible $B$ predictions.
Our beam search algorithm is very similar to the case of conventional left-to-right decoding, except that the decoding order might be arbitrary if we use confidence-based initial or refinement prediction methods.
As a result, extending different samples in the beam might lead to the same results so we need an additional deduplication step.
The time complexity with all the above components is $\mathcal{O}(M^2 B T)$, where $M$ is the maximal number of mask tokens, and $T$ is the maximal number of iteration.
\autoref{alg:overall} outlines the overall multi-token decoding algorithm.
The confidence-based decoding method takes 20 minutes to 2 hours on a Nvidia Geforce RTX 2080 Ti GPU depending on the number of facts of each language.

\begin{algorithm}[ht]
\small
\SetAlgoLined
\KwResult{The final sentence $\hat{\bm{s}}$.}
max number of mask tokens $M$, beam size $B$, max number of iteration $T$, an initial sentence $\bm{s}^{(0)}$\;
\For{number of mask tokens $m=1,...,M$}{
$\bm{s}_m^{(0)} \gets$ insert $m$ \mask tokens in $\bm{s}^{(0)}$\;
$\mathbf{S} \gets \{\bm{s}_m^{(0)}\}$\;
\For{iteration $t=1,...,T$}{
$\mathbf{S}^\prime \gets \phi$\;
\For{each sentence $\bm{s}_m^{(t-1)} \in \mathbf{S}$}{
 $\{\bm{s}_m^{(t,b)}\}_{b=1}^{B} \gets$ top $B$ predictions after an initial or refinement step\;
 $\mathbf{S}^\prime \gets \mathbf{S}^\prime \cup \{\bm{s}_m^{(t,b)}\}_{b=1}^{B}$
}
$\mathbf{S} \gets$ deduplicate and get the top $B$ from $\mathbf{S}^\prime$\;
}
}
$\hat{\bm{s}} \gets$ top one from $\mathbf{S}$\;
\caption{Multi-token decoding.}
\label{alg:overall}
\end{algorithm}

\begin{table}[ht]
\scriptsize
\begin{center}
\begin{tabular}{@{}p{3.8cm}@{\tinycol}l@{\tinycol}l@{}}
\toprule
\textbf{Prompts} & \textbf{Ind.} & \textbf{Best} \\
\midrule
The capital of India is \_. & Rajasthan & New Delhi \\
The capital of Auvergne is \_. & Lyon & Clermont-Ferrand \\
American League is part of \_. & the League & Major League Baseball \\
First Epistle to Timothy is part of \_. & Christianity & the New Testament \\
KGB is a legal term in \_. & KGB & the Soviet Union \\
Centers for Disease Control and & \multirow{2}{*}{CDC} & \multirow{2}{*}{the United States} \\
Prevention is a legal term in \_. \\
\bottomrule
\end{tabular}
\end{center}
\caption{Prediction results of M-BERT where the best-performing decoding method makes correct predictions while the independent prediction method does not.}
\label{tab:decode_en_case}
\end{table}

\section{Details of Pre-trained LMs}\label{ap:berts}

\begin{table*}[ht]
\small
\begin{center}
\begin{tabular}{lll}
\toprule
Model & Shortcut & Corpus \\
\midrule
\multicolumn{2}{c}{\emph{multilingual LMs}} \\
M-BERT & bert-base-multilingual-cased & Wikipedia \\
XLM & xlm-mlm-100-1280 & Wikipedia \\
XLM-R & xlm-roberta-base & CommonCrawl \\
\midrule
\multicolumn{2}{c}{\emph{monolingual LMs}} \\
BERT (en) & bert-base-cased & BooksCorpus, English Wikipedia \\
CamemBERT (fr) & camembert-base & French OSCAR$^\circ$ \\
BERTje (nl) & bert-base-dutch-cased & Dutch Wikipedia, Books, TwNC$^\star$, SoNaR-500$^\dagger$, Web news \\
BETO (es) & dccuchile/bert-base-spanish-wwm-cased & Spanish Wikipedia, Spanish OPUS$^\triangleleft$ \\
RuBERT (ru) & DeepPavlov/rubert-base-cased & Russian Wikipedia, news data\\
Chinese BERT (zh) & bert-base-chinese & Chinese Wikipedia \\
BERTurk (tr) & dbmdz/bert-base-turkish-cased & Turkish Wikipedia, Turkish OSCAR, Turkish OPUS, etc \\
GreekBERT (el) & nlpaueb/bert-base-greek-uncased-v1 & Greek Wikipedia, Greek Europarl$^\diamond$, Greek OSCAR \\
\bottomrule
\end{tabular}
\end{center}
\caption{Shortcut name of each multilingual/monolingual LM in HuggingFace's Transformers library, and their training copora. $^\circ$ The OSCAR corpus is extracted from the CommonCrawl corpus. $^\star$ TwNC is a multifaceted Dutch News Corpus. $^\dagger$ SoNaR-500 is a multi-genre Dutch reference corpus. $^\triangleleft$ OPUS is a translated text corpus from the web. $^\diamond$ Europarl is a corpus of parallel text.}
\label{tab:shortcut}
\end{table*}

LMs examined in this paper share similar architecture and pre-training setting as BERT \cite{devlin-etal-2019-bert} or RoBERTa \cite{liu-2019-roberta}, but are trained on different corpora.
We provide the shortcut name of each LM in the HuggingFace's Transformer library (\url{https://huggingface.co/transformers/pretrained_models.html}) and their training corpora in \autoref{tab:shortcut}, from which you can find more information.

\section{Detailed Experimental Results}

Detailed performance across LMs and languages and error cases in Spanish and Greek are shown in \autoref{tab:multilang} and \autoref{tab:err_more} respectively.

\begin{table*}[ht]
\small
\begin{center}
\begin{tabular}{l@{\tinycol}l@{\tinycol}l@{\tinycol}l@{\tinycol}r}
\toprule
\textbf{Type} & \textbf{Prompt} & \textbf{Prediction} & \textbf{Gold} & \textbf{Ratio} \\
\midrule
\emph{Correct} & Vilna y \_ son ciudades gemelas. & Minsk & Minsk & \emph{16.68} \\
\midrule
Repeating subjects & La capital de Bali es \_. & Bali & Denpasar & 24.62 \\
Wrong entities & John Goldschmidt es un \_ de profesiòn. & comerciant & director de cine & 29.07 \\
Non-informativeness & Lionel Heald fue educado en la Universidad de \_. & la Universidad & Charterhouse School & 9.81 \\
Type errors & Jänta å ja fue creada en \_. & 2005 & Suecia & 6.11 \\
Related concepts & Bas Heijne nació en \_. & el Reino de Holanda & Nimega & 1.67 \\
Unk & Tanaj consiste de \_. & :1.2 & Torá & 8.33 \\
False Negative & BMW S1000RR es producido por \_. & BMW & BMW Motorrad & 3.52 \\
Inflection & proteína de membrana es una subclase de \_. & proteínas & proteína & 0.19 \\
\midrule
\emph{Correct} & {\textgreekfont το Καμερούν βρίσκεται στην \_.} & {\textgreekfont Αφρική} & {\textgreekfont Αφρική} & \emph{12.02} \\
\midrule
Repeating subjects & {\textgreekfont η Λάσα ντε Σέλα δούλευε στην \_.} & {\textgreekfont Λάσα ντε Σέλα} & {\textgreekfont Μόντρεαλ} & 25.06 \\
Wrong entities & {\textgreekfont η Χένσελ ιδρύθηκε στην \_.} & {\textgreekfont Ιταλία} &{\textgreekfont Κάσσελ} & 18.74 \\
Non-informativeness & {\textgreekfont ο Πωλ Καρνό δουλεύει στο \_.} & {\textgreekfont χωριό} &  {\textgreekfont Πανεπιστήμιο ραρισιού} & 26.78 \\
Related concepts & {\textgreekfont οι} The Kooks {\textgreekfont ιδρύθηκαν στην \_.} & {\textgreekfont Αγγλία} & {\textgreekfont Μπράιτον} & 1.91 \\
Unk & {\textgreekfont ο Ραβί Σανκάρ παίζει \_.} & {\textgreekfont π} & {\textgreekfont Σιτάρ} & 11.67 \\
False Negative & {\textgreekfont το} Disneyland {\textgreekfont ανήκει στο \_.} & Walt Disney & the Walt Disney Company  & 3.06 \\
Inflection & {\textgreekfont ο Χριστός είναι μέρος του \_.} & {\textgreekfont Χριστός} & {\textgreekfont Χριστού} & 0.77 \\
\bottomrule
\end{tabular}
\end{center}
 \caption{Error cases of M-BERT in Spanish and Greek (\%).}
\label{tab:err_more}
\end{table*}

\begin{table*}[ht]
\small
\begin{center}
\begin{tabular}{l@{\tinycol}l@{\tinycol}l|r@{\tinycol}r@{\tinycol}r@{\tinycol}r|r@{\tinycol}r@{\tinycol}r@{\tinycol}r@{\tinycol}r@{\tinycol}r@{\tinycol}r@{\tinycol}r@{\tinycol}r@{\tinycol}r}
\toprule
\textbf{Model} & \textbf{Decoding} & \textbf{Part} & \textbf{en} & \textbf{fr} & \textbf{nl} & \textbf{es} & \textbf{ru} & \textbf{zh} & \textbf{he} & \textbf{tr} & \textbf{ko} & \textbf{vi} & \textbf{el} & \textbf{mr} & \textbf{yo} \\
\midrule
\multirow{6}{*}{M-BERT} & \multirow{3}{*}{Ind.} & all & 13.57 & 10.21 & \textbf{12.42} & \textbf{14.30} & 1.87 & 2.50 & 2.70 & 2.00 & 4.08 & 8.34 & 4.46 & 2.76 & \textbf{3.44} \\
 &  & single & 22.40 & 19.07 & \textbf{25.21} & \textbf{24.25} & 4.58 & 9.61 & 7.43 & 4.50 & \textbf{21.14} & \textbf{17.69} & 21.11 & 12.11 & \textbf{5.15} \\
 &  & multi & 5.57 & 3.92 & 4.42 & 4.90 & 0.96 & 2.22 & 2.56 & 1.03 & 1.61 & 2.91 & 2.16 & 2.18 & \textbf{3.29} \\
\cmidrule{2-17}
 & \multirow{3}{*}{Conf.} & all & 12.00 & 6.30 & 8.55 & 7.47 & 2.54 & 6.62 & 2.92 & 2.08 & 4.70 & 9.20 & 6.77 & \textbf{3.46} & 3.21 \\
 &  & single & 12.91 & 7.77 & 12.20 & 9.13 & 3.65 & 5.21 & 4.33 & 4.34 & 16.15 & 14.60 & 13.69 & 8.99 & 3.87 \\
 &  & multi & \textbf{10.08} & 4.78 & \textbf{5.22} & 5.11 & 1.86 & 6.49 & \textbf{2.90} & 1.19 & 2.88 & 5.22 & 5.72 & \textbf{3.07} & 3.06 \\
\midrule
\multirow{6}{*}{XLM} & \multirow{3}{*}{Ind.} & all & 9.03 & 7.44 & 7.53 & 7.40 & 2.29 & 5.83 & 2.79 & 1.59 & 5.33 & 6.86 & 7.10 & 1.26 & - \\
 &  & single & 20.74 & 16.58 & 18.38 & 16.44 & 7.62 & \textbf{17.12} & 11.58 & 5.53 & 13.28 & 12.12 & 18.03 & 12.62 & - \\
 &  & multi & 4.75 & 4.03 & 3.00 & 3.40 & 1.40 & 2.57 & 1.82 & 0.50 & 3.24 & 3.93 & 5.16 & 0.10 & - \\
\cmidrule{2-17}
 & \multirow{3}{*}{Conf.} & all & 5.30 & 4.13 & 4.46 & 3.18 & 2.14 & 3.40 & 1.93 & 1.85 & 5.23 & 6.26 & \textbf{7.56} & 1.48 & - \\
 &  & single & 8.79 & 6.14 & 6.48 & 4.18 & 3.61 & 10.44 & 5.97 & 4.89 & 11.15 & 8.98 & 13.86 & 9.76 & - \\
 &  & multi & 5.63 & 3.56 & 4.06 & 3.09 & 2.01 & 1.38 & 1.71 & 1.06 & \textbf{3.82} & 4.38 & \textbf{6.50} & 0.42 & - \\
\midrule
\multirow{6}{*}{XLM-R} & \multirow{3}{*}{Ind.} & all & 8.19 & 4.70 & 4.42 & 6.50 & 5.26 & 4.63 & 2.47 & 3.09 & 5.11 & 8.52 & 6.28 & 2.71 & - \\
 &  & single & 15.21 & 11.29 & 10.95 & 13.37 & \textbf{14.41} & 11.85 & \textbf{12.34} & 4.04 & 16.71 & 14.22 & \textbf{27.33} & \textbf{19.47} & - \\
 &  & multi & 3.32 & 2.34 & 2.58 & 3.29 & 3.77 & 4.49 & 2.18 & \textbf{2.49} & 2.61 & 5.12 & 2.94 & 1.07 & - \\
\cmidrule{2-17}
 & \multirow{3}{*}{Conf.} & all & 4.43 & 2.90 & 2.67 & 4.33 & 5.53 & 5.30 & \textbf{2.99} & 2.95 & \textbf{5.64} & \textbf{9.51} & 7.25 & 3.36 & - \\
 &  & single & 5.19 & 4.38 & 3.57 & 4.93 & 14.15 & 11.79 & 11.42 & 3.93 & 15.88 & 12.56 & 25.60 & 18.85 & - \\
 &  & multi & 3.86 & 2.33 & 2.70 & 4.17 & 4.12 & 5.17 & 2.73 & 2.43 & 3.44 & \textbf{6.97} & 4.29 & 1.97 & - \\
\midrule
\multirow{6}{*}{Specific} & \multirow{3}{*}{Ind.} & all & \textbf{17.92} & \textbf{10.36} & 9.84 & 10.94 & 6.77 & 5.47 & - & \textbf{3.36} & - & - & 3.00 & - & - \\
 &  & single & \textbf{31.21} & \textbf{20.30} & 19.22 & 19.07 & 9.64 & 3.55 & - & \textbf{5.88} & - & - & 5.53 & - & - \\
 &  & multi & 5.88 & \textbf{4.88} & 3.40 & 6.10 & 5.50 & 5.18 & - & 2.29 & - & - & 0.92 & - & - \\
\cmidrule{2-17}
 & \multirow{3}{*}{Conf.} & all & 10.53 & 6.20 & 5.18 & 6.07 & \textbf{6.80} & \textbf{10.07} & - & 3.13 & - & - & 2.49 & - & - \\
 &  & single & 19.01 & 15.50 & 8.21 & 5.22 & 9.22 & 3.04 & - & 5.56 & - & - & 4.08 & - & - \\
 &  & multi & 3.44 & 3.09 & 3.06 & \textbf{6.40} & \textbf{5.59} & \textbf{9.80} & - & 2.15 & - & - & 1.35 & - & - \\
\toprule 
\textbf{Model} & \textbf{Decoding} & \textbf{Part} &  & &  & &  
\textbf{ja} & \textbf{hu} & \textbf{bn} & \textbf{ceb} & \textbf{war} & \textbf{tl} & \textbf{sw} & \textbf{pa} & \textbf{mg} & \textbf{ilo} \\
\midrule
\multirow{6}{*}{M-BERT} & \multirow{3}{*}{Ind.} & all & & & & & 0.85 & 2.54 & 1.33 & 3.93 & 2.29 & \textbf{5.41} & \textbf{6.24} & 1.91 & \textbf{3.36} & \textbf{1.82} \\
 &  & single & & & & & 7.13 & \textbf{8.31} & \textbf{2.39} & \textbf{7.13} & 4.42 & \textbf{10.12} & \textbf{10.00} & 4.35 & \textbf{4.36} & \textbf{3.06} \\
 &  & multi & & & & & 0.48 & 0.62 & 1.12 & 0.23 & 0.42 & 0.64 & 2.25 & 1.48 & 3.27 & 0.19 \\
\cmidrule{2-17} 
 & \multirow{3}{*}{Conf.} & all & & & & & 1.51 & \textbf{3.16} & \textbf{1.51} & 3.94 & 2.11 & 4.62 & 6.02 & \textbf{2.56} & 3.27 & 1.70 \\
 &  & single & & & & & 6.50 & 7.85 & 1.52 & 6.30 & 3.73 & 7.80 & 8.42 & 3.80 & 3.40 & 2.41 \\
 &  & multi & & & & & 1.21 & \textbf{1.68} & \textbf{1.34} & 0.64 & 0.69 & 1.25 & \textbf{3.60} & \textbf{2.30} & \textbf{3.52} & 0.24 \\
\midrule
\multirow{6}{*}{XLM} & \multirow{3}{*}{Ind.} & all & & & & & 5.77 & 1.56 & 0.10 & \textbf{5.39} & \textbf{3.29} & 4.36 & 5.90 & - & - & 0.13 \\
 &  & single & & & & & \textbf{24.95} & 6.71 & 1.13 & 6.98 & \textbf{5.35} & 7.35 & 8.60 & - & - & 0.43 \\
 &  & multi & & & & & 3.04 & 0.60 & 0.00 & \textbf{2.15} & \textbf{1.83} & 1.36 & 2.18 & - & - & 0.00 \\
\cmidrule{2-17}
 & \multirow{3}{*}{Conf.} & all & & & & & \textbf{5.95} & 1.87 & 0.06 & 4.67 & 1.57 & 2.25 & 4.19 & - & - & 0.04 \\
 &  & single & & & & & 18.60 & 5.49 & 0.81 & 4.88 & 2.17 & 3.53 & 5.90 & - & - & 0.07 \\
 &  & multi & & & & & \textbf{4.24} & 1.34 & 0.00 & 2.11 & 1.08 & 1.11 & 2.28 & - & - & 0.00 \\
\midrule
\multirow{6}{*}{XLM-R} & \multirow{3}{*}{Ind.} & all & & & & & 2.30 & 0.86 & 0.07 & 1.35 & 1.15 & 2.80 & 3.66 & 0.23 & 1.94 & 0.11 \\
 &  & single & & & & & 9.23 & 2.22 & 0.00 & 1.73 & 1.32 & 5.05 & 5.57 & 5.75 & 3.70 & 0.39 \\
 &  & multi & & & & & 2.07 & 0.24 & 0.07 & 1.03 & 1.08 & \textbf{1.42} & 1.91 & 0.00 & 1.61 & 0.02 \\
\cmidrule{2-17}
 & \multirow{3}{*}{Conf.} & all & & & & & 4.41 & 0.86 & 0.09 & 1.22 & 1.14 & 2.33 & 2.86 & 0.58 & 1.76 & 0.51 \\
 &  & single & & & & & 8.82 & 2.02 & 0.00 & 1.39 & 1.29 & 4.25 & 4.34 & \textbf{5.75} & 3.49 & 0.39 \\
 &  & multi & & & & & 4.21 & 0.31 & 0.10 & 0.99 & 1.07 & 1.28 & 1.85 & 0.36 & 1.45 & \textbf{0.52} \\
\bottomrule
\end{tabular}
\end{center}
\caption{Accuracy on different languages using different LMs (\%). We use $M=5$ mask tokens for en, fr, nl es, vi (on the left) and $M=10$ mask tokens for the other languages on the right. Best results for each language-part combination are in bold. ``-'' denotes missing/unsupported models.}
\label{tab:multilang}
\end{table*}

\end{document}